\pdfoutput=1

\documentclass{article}

\usepackage{arxiv}
\usepackage[utf8]{inputenc}
\DeclareUnicodeCharacter{2212}{ }

\usepackage[T1]{fontenc}    
\usepackage{url}            
\usepackage{booktabs}       
\usepackage{amsfonts}       
\usepackage{nicefrac}       
\usepackage{microtype}      
\usepackage{graphicx}
\usepackage{doi}
\usepackage{bbding}

\usepackage{amsmath,amssymb,amsfonts}
\usepackage{hyperref}
\usepackage[binary-units=true]{siunitx}
\usepackage{cases}
\usepackage{booktabs}
\usepackage[inline]{enumitem}
\usepackage{makecell}
\usepackage{subcaption}
\usepackage[table]{xcolor}

\usepackage{tikz}
\usepackage{color}
\usetikzlibrary{shapes,arrows}

\usepackage{xcolor}
\hypersetup{
    hidelinks,
    linkcolor={red!50!black},
    citecolor={blue!50!black},
    urlcolor={blue!80!black}
}

\DeclareGraphicsExtensions{.pdf,.jpeg,.png}

\newcommand{\M}[1]{\mathbf{\MakeUppercase{#1}}}
 
\newcommand{\V}[1]{\bar{\boldsymbol{\mathbf{\MakeLowercase{#1}}}}}
\newcommand{\NORM}[1]{\left\Vert{#1}\right\Vert}
\newcommand{\Vf}[4]{{}^{\F{#1}}\bar{\boldsymbol{\mathbf{\MakeLowercase{#2}}}}^{\F{#3}}_{\F{#4}}} 
\newcommand{\Vfdot}[4]{{}^{\F{#1}}\dot{\bar{\boldsymbol{\mathbf{\MakeLowercase{#2}}}}}^{\F{#3}}_{\F{#4}}}

\newcommand{\Qf}[4]{{}^{\F{#1}}\boldsymbol{\mathbf{\MakeLowercase{#2}}}^{\F{#3}}_{\F{#4}}}

\newcommand{\F}[1]{\mathtt{#1}}

\newcommand{\SET}[1]{\mathit{#1}}

\newcommand{\GROUP}[1]{\mathbb{#1}}

\newcommand{\TR}[3]{{}^{\F{#2}}\M{#1}_{\F{#3}}}

\newcommand{\CROSS}[1]{\left[#1\right]_{\times}}

\newcommand{\Q}[2]{\left[\mathbf{#1}\right]_{\textsl{#2}}}
\newcommand{\Qx}[1]{\boldsymbol{\mathbf{#1}}}

\newcommand{\best}[1]{{\cellcolor[gray]{0.8}} #1}

\title{A Hybrid Modelling Approach for Aerial Manipulators}

\author{ \href{https://orcid.org/0000-0001-9220-6679}{\includegraphics[scale=0.06]{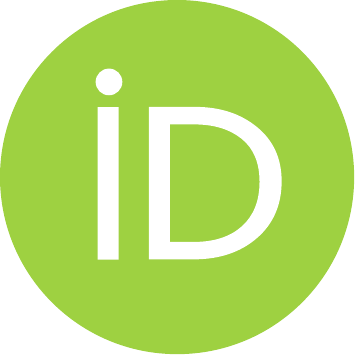}\hspace{1mm}Paul Kremer}\thanks{University of Luxembourg,
    Faculty of Science, Technology and Medicine (FSTM),
	2, Av. de l'Université, L-4365 Esch-sur-Alzette, Luxembourg} \hspace{1mm} \Envelope \\
	\texttt{p.kremer@uni.lu}\\
	\And
	\href{https://orcid.org/0000-0001-5018-0925}{\includegraphics[scale=0.06]{orcid.pdf}\hspace{1mm}Jose Luis Sanchez-Lopez}\thanks{Interdisciplinary Center for Security, Reliability and Trust (SnT),
	University of Luxembourg,
	29, Av. J.F. Kennedy, L-1855 Luxembourg} \\
    \texttt{joseluis.sanchezlopez@uni.lu} \\
  	\And
	\href{https://orcid.org/0000-0002-9600-8386}{\includegraphics[scale=0.06]{orcid.pdf}\hspace{1mm}Holger Voos}\footnotemark[1] \hspace{0.5mm} \footnotemark[2]\\
    \texttt{holger.voos@uni.lu} \\
}

\begin{document}
\maketitle

\begin{abstract}
    Aerial manipulators (AM) exhibit particularly challenging, non-linear dynamics; the UAV and the manipulator it is carrying form a tightly coupled dynamic system, mutually impacting each other. The mathematical model describing these dynamics forms the core of many solutions in non-linear control and deep reinforcement learning.
    Traditionally, the formulation of the dynamics involves Euler angle parametrization in the Lagrangian framework or quaternion parametrization in the Newton-Euler framework. The former has the disadvantage of giving birth to singularities and the latter of being algorithmically complex. This work presents a hybrid solution, combining the benefits of both, namely a quaternion approach leveraging the Lagrangian framework, connecting the singularity-free parameterization with the algorithmic simplicity of the Lagrangian approach. We do so by offering detailed insights into the kinematic modeling process and the formulation of the dynamics of a general aerial manipulator.
    The obtained dynamics model is validated experimentally against a real-time physics engine. A practical application of the obtained dynamics model is shown in the context of a computed torque feedback controller (feedback linearization), where we analyze its real-time capability with increasingly complex models.
\end{abstract}

\subsection*{License}
For the purpose of Open Access, the author has applied a CC-BY-4.0 public copyright license to any Author Accepted Manuscript version arising from this submission.

\section{Introduction}
The field of aerial manipulation has seen a lot of interest in recent years \cite{Ollero2019, Mohiuddin2020} with solutions being developed ranging from bridge inspection \cite{Bartelds2016, Ikeda2017}, to maintenance \cite{Korpela2014b}, to assembly \cite{Jimenez-Cano2013}, various pick and place tasks \cite{Lippiello2012a, Heredia2014b, Garimella2015}, slung load transportation \cite{Bernard2009}, cooperative transportation \cite{Michael2011}, and even drilling and screwing \cite{Ding2021}. UAVs are thus becoming increasingly aware of their surroundings, allowing them to autonomously navigate \cite{Sanchez-Lopez2020} and interact with the environment.

\begin{figure}
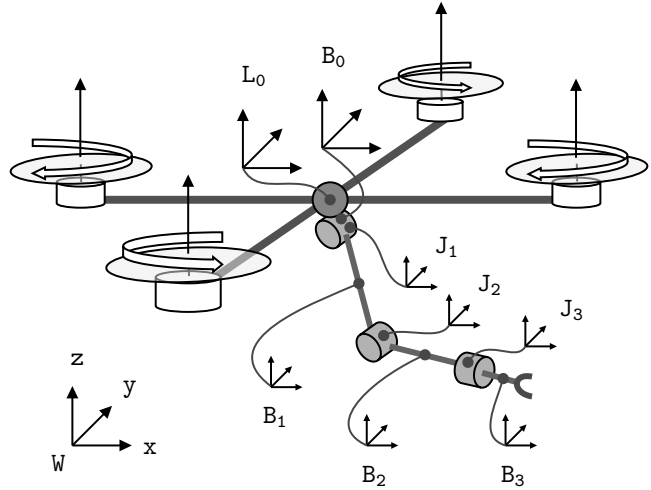

    \centering
    \include{figure_muav}
    \caption{Illustration of a typical aerial manipulator; a quadcopter with a 3-DOF serial link manipulator. Indicated are the frames of reference for the joints $\F{J}_i$, the inertial (body) frames $\F{B}_i$ and forces of the propulsion devices for thrust $\V{f}_{thr}$ and drag $\V{f}_{drag}$.}\label{fig:problem-illutration}
\end{figure}

Some aerial manipulation tasks can be performed with relatively simple static claws \cite{Mellinger2011} with only minimal impact on the carrying platform. However, more complex tasks require the additional degrees of freedom offered by serial link manipulators \cite{Suarez2017, Korpela2014b, Orsag2013, Lippiello2012a}. A typical AM is shown in figure \ref{fig:problem-illutration}. Those manipulators are often relatively heavy and have a substantial dynamic impact on their carrying, generally, under-actuated flying platform. This impact stems from multiple sources, such as the reaction forces and torques related to the movement of the manipulator, or the shifting center of gravity related to the changing physical configuration of the manipulator, and by contact forces due to the direct interaction with the environment.

A mathematical model describing the tightly coupled dynamics of such systems is often key to performing complicated aerial manipulation tasks. The derivation of the kinematics and dynamics model is a well-studied subject in classical robotics, falling under the category of mobile robotics \cite{BrunoSiciliano2008}, where the floating base is modeled via a 6-DOF joint. In aerial manipulation, however, the parametrization of rotational DOF often employs Euler angles \cite{Jimenez-Cano2013, Kim2013, Suarez2017, Jiao2019}, which gives birth to singularities, or complex dual number formulations (dual quaternion) \cite{Abaunza2016a, Abaunza2017}, which can be hard to implement.

As tasks in aerial manipulation gain in complexity, Euler angles (resp. any three-angle representation of orientations) become a burden as they exhibit gimbal lock \cite{Hemingway2018} in certain conditions (fig. \ref{fig:quaterion_euler_motivation}). Traditional UAVs generally operate at a safe distance from those conditions, but they can still exhibit them during certain maneuvers, e.g., by flipping them about their roll axis or as a consequence of interaction with their surroundings. Therefore, it is important for the control designer to have a dynamics model that stays valid under any circumstances. To that end, we propose a hybrid solution to the dynamics modeling: by combining unit quaternion parameterization with the simplicity of the Lagrangian framework, a singularity-free, dynamics model is obtained without resorting to the complexity of dual number formulations resp. the algorithmic complexity of Newton-based methods. Our approach is kept relatively general with the goal to cover most AM configurations (even tilting propeller configurations - to some extent), and the different steps are described in detail to facilitate practical implementations.

\subsection{Related Work}
The dynamics of a multi-body system are commonly derived using either Newton or energy-based methods. The recursive Newton-Euler algorithm (RNEA) and the composite-rigid-body algorithm (CRBA) \cite{BrunoSiciliano2008} are very popular choices of Newton-based methods. The Euler-Lagrange equations are conceptually much simpler and thus easier to implement. However, Newton-based methods often yield more compact solutions \cite{Tassora2001} at the expense of being algorithmically more complex.

Currently, it is common practice to combine quaternion attitude representation with a complex, Newton-based method and Euler attitude representation with the simpler Lagrangian formulation. The authors of \cite{Kim2013, Jiao2019} derived the coupled model of an AM with a 2-DOF manipulator using Euler angles and Lagrangian mechanics. The same approach was chosen for a dual 5-DOF arm configuration in \cite{Suarez2017} and a 3-DOF manipulator in \cite{Emami2021}. A dual-quaternion approach employing the Euler-Lagrange formalism has been chosen in \cite{Abaunza2017}. In \cite{Orsag2013}, the authors chose to derive the dynamics of three 2-DOF manipulators using the Newton-Euler method. The authors of \cite{Fanni2017} followed the same approach, with a focus on inverse kinematics with holonomic constraints. In \cite{Jimenez-Cano2013}, the dynamics were derived using the Newton-Euler method and Euler angles. In \cite{Wang2019}, the authors presented an approach within the Lagrangian framework (employing Euler angles) that handles constraints applied to the end-effector by introducing an additional constraint term to the dynamics equation. In this work, the coupled dynamics, using singularity-free quaternion representation, is derived in detail using the simpler Euler-Lagrange equations, thus offering an appealing alternative (resp. middle ground) to the state of the art.

This work primarily focuses on the dynamics model; therefore, only a very simple PD controller with a bias correction term gets introduced because it does not hide model imperfections. The state of the art with regard to model-based control techniques can be found in \cite{Nascimento2019}. Some highlights include the robust, adaptive sliding mode controller introduced in \cite{Kim2013}, the variable parameter integral backstepping controller in \cite{Jimenez-Cano2013, Heredia2014b}, the passivity based controller in \cite{Acosta2014}, the disturbance observer-based control in \cite{Fanni2017}, the hybrid position and force controller respecting constraint conditions presented in \cite{Wang2019}, and the model reference adaptive control in \cite{Ali2020}. The authors of \cite{Liang2021} specifically addressed the problem of model uncertainty and unknown disturbances by employing a nonlinear disturbance observer in conjunction with a robust prescribed performance controller. All of those control schemes have the dynamics model of their controlling system at their core whilst being robust towards modeling imperfections.

\subsection{Contributions}
The main contribution of this work is a general-purpose, user-friendly, yet powerful dynamics framework for aerial manipulation that yields a singularity-free, closed-form dynamics model of an aerial manipulator. Although we do not target tilting propeller resp. fully actuated UAV configurations in this work, the results presented herein are general enough to be applicable to those configurations as well (assuming that the reaction torque of the tiling mechanism and its impact on the center of gravity are negligible).

The contributions of this work are thus stated as follows:
\begin{enumerate*}
    \item Providing a hybrid approach with regard to the dynamics modeling of aerial manipulators, by combining quaternion parametrization and Lagrangian mechanics. This is contrary to what most authors do in this field, namely using either Euler angle parameterization in the Lagrangian framework, or quaternion parameterization in the Newton-Euler framework. The approach presented here leads to a relatively generic, singularity-free, coupled dynamics model. Implementing our approach requires significantly less coding than Newton-based methods, thanks to the algorithmic simplicity of the Lagrangian framework. In fact, the bulk of the work consists in implementing the simple kinematic equations, contrary to the Newton-Euler algorithm, which also involves expressing the respective force and torque equations in a recursive manner \cite{BrunoSiciliano2008}.
    \item The derivation of the kinematics resp. the dynamic model often comes up short in many publications, where the focus is often on control rather than the modeling. This perceived lack of modeling methodology is addressed herein, showing all the required steps and thus providing insights into the normally hidden process, lowing the barrier to entry in the field of aerial manipulation.
    \item The introduction of a general-purpose dynamics framework for aerial manipulators based on a URDF~\cite{ROS} description of the system.
    \item By making use of the Lagrangian mechanics framework and by refraining from using dual quaternions (dual numbers), in combination with the mapping of all quaternion operations to matrix-vector products, the results presented here are particularly well suited to be implemented with common computer algebra systems (CAS).
    \item Two applications of the model are shown. First, in a simulation resp. validation context where the obtained model is validated against an identical reference model simulated in a realtime physics engine. And second, as part of an adaptation of a computed torque controller adopted from classical robotics to work in an aerial manipulation context.
    \item Performance figures with regard to the algebraic complexity and the real-time capability are provided for two typical scenarios, namely onboard resp. off-board control using increasingly complex AM. Furthermore, we compare the complexity of the resulting dynamics equations obtained from different methods.
\end{enumerate*}

\subsection{Structure}
This work is organized as follows. We start with the kinematics of a single body in the inertial frame, followed by the kinematic equations of an open kinematic chain manipulator with regard to its own base frame. Then, by combining those results, the base of the robot is turned into a floating platform, providing an additional 6-DOF, which results in the complete kinematic equations of the manipulating system in the world frame. The kinematic equations are then fed into  the Lagrangian framework to obtain the dynamics equations. It follows the non-trivial mapping between body forces and generalized forces resp. propulsion forces and body forces. Finally, a general-purpose holonomic constraint solver is applied to the unconstrained system to impose the quaternion unit constraint via a constraint force. The obtained dynamics model is then validated in simulation against a real-time physics engine. As the last application, a stabilizing controller is introduced that uses the developed dynamics model to compensate for the nonlinearities present in the AM. This work is concluded by presenting performance figures concerning the algebraic complexity and the real-time capability of the model within the control context. A short introduction to (unit) quaternions is provided in \ref{sec:quaternion-fundamentals}.

The nomenclature is as stated in table \ref{tab:nomenclature}.

\begin{table}
    \centering
    \caption{Notation, frequently used symbols}\label{tab:nomenclature}
    \begin{tabular}{ll}
        \toprule
        Symbol                 & Definition                                                   \\\toprule
        $\V{p}$                & Vector $\V{p}$                                               \\\midrule
        $\F{W}$                & Inertial frame                                               \\\midrule
        $\F{B_i}$              & Body frame $i$                                               \\\midrule
        $\F{J_i}$              & Joint frame $i$                                              \\\midrule
        $\F{L_i}$              & Link frame $i$                                               \\\midrule
        $\Vf{A}{n}{}{B}$       & A normal vector situated in $\F{B}$ expressed in $\F{A}$     \\\midrule
        $\Vf{A}{p}{C}{D}$      & A vector that spans from $\F{C}$ to $\F{D}$, in $\F{A}$      \\\midrule
        $\Vf{A}{\omega}{C}{D}$ & Angular velocity of $\F{D}$, relative to $\F{C}$, in $\F{A}$ \\\midrule
        $\Vf{A}{v}{C}{D}$      & Linear velocity of $\F{D}$, relative to $\F{C}$, in $\F{A}$  \\\midrule
        $\M{A}$                & Matrix $\M{A}$                                               \\\midrule
        $\TR{R}{W}{B}$         & Rotation matrix mapping from $\F{B}$ to $\F{W}$              \\\midrule
        $\TR{A}{W}{B}$         & Homogeneous transformation mapping from $\F{B}$ to $\F{W}$   \\\midrule
        $\Qx{q}$               & Quaternion                                                   \\\midrule
        $N_J$                  & Number of joints                                             \\\midrule
        $N_B$                  & Number of bodies                                             \\\midrule
        $N_L$                  & Number of links                                              \\\midrule
        $N_x$                  & Number of coordinates                                        \\\bottomrule
    \end{tabular}
\end{table}

\section{Kinematics}
The following section derives the forward kinematics of the combined (coupled) system formed by the manipulator and its carrying, flying platform. Commonly, this is done by introducing a 6-DOF joint between the world frame and the first body. In this work, however, it is more convenient first to describe the kinematics of the two subsystems (drone and manipulator) and then, as the last step, merge both systems together. The drone then acts as the 6-DOF joint.

\subsection{UAV Body Kinematics}
\begin{figure}
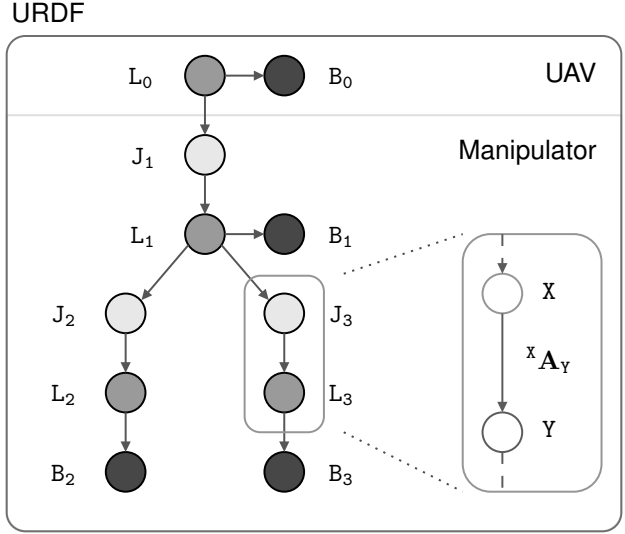

    \centering
    \include{figure_kinematic_tree}
    \caption{Example of a URDF-compliant kinematic tree of a branched kinematic chain. The link and joint frames are annotated as $\F{L}_i$ resp. $\F{J}_i$. The body frame $\F{B_i}$ has an associated mass and moment of inertia. It marks the center of mass of its parent link $\F{J}_i$. $\F{L_0}$ marks the base link of the robot (namely the UAV). The edges between the nodes (see detail) represent coordinate transformations between two frames denoted by $\TR{A}{X}{Y}$.}\label{fig:transforms}
\end{figure}

Given a UAV, represented by base link $\F{L_0}$ and the body $\F{B_0}$ in the kinematic chain as depicted in figure \ref{fig:transforms}. Its orientation is described by the quaternion $\Qx{q}=\left(q_w, \V{q}_{v}\right) \in \SET{S}^3$, and its position in the world frame $\F{W}$ is given by the vector
\begin{equation}
    \Vf{W}{p}{W}{L_0}=\left[x,y,z\right] \in \GROUP{R}^3.
\end{equation}
The base link's angular velocity $\Vf{W}{\omega}{W}{L_0} \in \GROUP{R}^3$, expressed in $\F{W}$, relative to $\F{W}$, is tied to the quaternion $\Qx{q}$ via
\begin{equation}
    \left(0, \Vf{W}{\omega}{W}{L_0} \right) = 2 \Qx{\dot{q}} \otimes \Qx{q}^*.
    \label{eq:ww}
\end{equation}
Which can be written in matrix form using \eqref{eq:qr}
\begin{equation}
    \Qf{W}{\omega}{W}{L_0} = 2 \Q{q}{R}^\intercal \Qx{\dot{q}}.
    \label{eq:body-angular-velocity-inertial}
\end{equation}
Similarly, the base link's angular velocity expressed in $\F{L_0}$, relative to $\F{W}$ can be written as
\begin{equation}
    \left(0, \Vf{L_0}{\omega}{W}{L_0} \right) = 2 \Qx{q}^* \otimes \Qx{\dot{q}},
    \label{eq:wb}
\end{equation}
respectively with the help of \eqref{eq:ql}
\begin{equation}
    \Qf{L_0}{\omega}{W}{L_0} = 2 \Q{q}{L}^\intercal \Qx{\dot{q}}.
    \label{eq:body-angular-velocity-body}
\end{equation}
Since $\Qf{W}{\omega}{W}{L_0}$ and $\Qf{L_0}{\omega}{W}{L_0}$ are pure quaternions, their vector part can directly be obtained by dropping the first row from~\eqref{eq:body-angular-velocity-inertial} resp.~\eqref{eq:body-angular-velocity-body} yielding the two orthogonal mapping matrices $\M{E}$ and $\M{G}$ defined by
\begin{align}
    \Vf{W}{\omega}{W}{L_0} = 2 \underbrace{\begin{bmatrix}
            -\V{q}_v & q_w \M{1}_3 + \CROSS{\V{q}_v}
        \end{bmatrix}}_{\M{E} \in \GROUP{R}^{3\times4}} \Qx{\dot{q}}
    \label{eq:E}
\end{align}
resp.
\begin{align}
    \Vf{L_0}{\omega}{W}{L_0} = 2 \underbrace{\begin{bmatrix}
            -\V{q}_v & q_w \M{1}_3 - \CROSS{\V{q}_v}
        \end{bmatrix}}_{\M{G} \in \GROUP{R}^{3\times4}} \Qx{\dot{q}}.
    \label{eq:G}
\end{align}
Solving~\eqref{eq:G} for $\Qx{\dot{q}}$ and replacing it into~\eqref{eq:E} reveals the relation between $\Qf{W}{\omega}{W}{L_0}$ and $\Qf{L_0}{\omega}{W}{L_0}$:
\begin{equation}
    \Vf{W}{\omega}{W}{L_0} = \underbrace{\M{E} \M{G}^\intercal}_{\TR{R}{W}{L_0}} \Vf{L_0}{\omega}{W}{L_0}, \label{eq:quaternion-rotmat}
\end{equation}
where $\TR{R}{W}{L_0} \in \SET{SO}\left(3\right)$ is the orthogonal rotation matrix associated with $\Qx{q}$.

\subsection{Kinematics of the Manipulator}
In the following section, the forward kinematics of an open-chain manipulator is developed. Although the base link of the manipulator is represented by the UAV, for the development in this section, the base link $\F{L_0}$ is regarded as static and all quantities of motion will be expressed in and with regard to $\F{L_0}$ (figure \ref{fig:transforms}). The development largely follows the procedures found in classical robotics literature \cite{Murray1994},~\cite{Craig2005},~\cite{BrunoSiciliano2008} and~\cite{Siciliano2009} with some changes to the kinematic tree that are required to interface with the Universal Robot Description Format specifications seamlessly.

A generic high-level description of a robotic arm can be stated in form of a URDF, providing all required kinematic and physical properties of a robotic manipulator. Its conventions concerning reference frames, links, joints and parents have been adopted here to facilitate the translation from an AM given as a URDF file to its system dynamics.

The URDF structure is standardized and describes an (open) kinematic chain formed by various types of nodes. Nodes are defined as either joints $\F{J_i}$, links $\F{L_i}$ or bodies $\F{B_i}$ (fig.~\ref{fig:transforms}), and are related to each other by homogeneous transformations such as
\begin{equation}
    \TR{A}{X}{Y} = \begin{bmatrix}
        \TR{R}{X}{Y} & \Vf{X}{p}{X}{Y} \\
        \V{0}        & 1
    \end{bmatrix} \in \SET{SE(3)}.
    \label{eq:hom-transf}
\end{equation}

The key aspects of the URDF are summarized as follows (refer to ~\cite{ROS} for further details):
\begin{itemize}
    \item The kinematic chain starts with the base link $\F{L_0}$
    \item Each link can be the parent of multiple joints, thus creating branched structures
    \item Each joint is the parent of the single link
    \item Each link has a (rigid) body associated to it. Its frame $\F{B_i}$ defines the center of mass (CM) of the body having the mass $m_i$ and the moment of inertia $\M{\Phi}_i$.
    \item The transformation (edge) between two nodes is defined by a homogeneous transformation $\TR{A}{X}{Y}$
\end{itemize}
Those physical properties $m_i$, $\M{\Phi}_i$ and $\TR{A}{X}{Y}$ are best obtained directly from CAD data.

\begin{figure}
    \centering
    \tikzset{every picture/.style={line width=0.75pt}} 

\begin{tikzpicture}[x=0.75pt,y=0.75pt,yscale=-1,xscale=1]

\draw [color={rgb, 255:red, 208; green, 208; blue, 208 }  ,draw opacity=1 ][line width=2.25]    (285.49,35.23) -- (303.82,104.98) ;
\draw  [color={rgb, 255:red, 74; green, 74; blue, 74 }  ,draw opacity=1 ][fill={rgb, 255:red, 222; green, 222; blue, 222 }  ,fill opacity=1 ] (299.08,108.54) -- (307.78,102.3) .. controls (310.18,100.58) and (314.33,102.23) .. (317.04,105.98) .. controls (319.75,109.74) and (320.01,114.18) .. (317.61,115.9) -- (308.91,122.13) .. controls (306.51,123.85) and (302.36,122.2) .. (299.65,118.45) .. controls (296.93,114.7) and (296.68,110.26) .. (299.08,108.54) .. controls (301.48,106.82) and (305.63,108.46) .. (308.34,112.22) .. controls (311.06,115.97) and (311.31,120.41) .. (308.91,122.13) ;
\draw [color={rgb, 255:red, 148; green, 93; blue, 78 }  ,draw opacity=1 ][line width=0.75]    (272.45,141.92) -- (303.47,116.43) ;
\draw [shift={(303.47,116.43)}, rotate = 500.58] [color={rgb, 255:red, 148; green, 93; blue, 78 }  ,draw opacity=1 ][line width=0.75]    (7.06,-1.32) .. controls (5.47,-0.56) and (4.01,-0.12) .. (2.68,0) .. controls (4.01,0.12) and (5.47,0.56) .. (7.06,1.32)(4.37,-1.32) .. controls (2.78,-0.56) and (1.32,-0.12) .. (0,0) .. controls (1.32,0.12) and (2.78,0.56) .. (4.37,1.32)   ;
\draw [color={rgb, 255:red, 208; green, 208; blue, 208 }  ,draw opacity=1 ][line width=2.25]    (314.63,117) -- (350,138) ;
\draw  [color={rgb, 255:red, 74; green, 74; blue, 74 }  ,draw opacity=1 ][fill={rgb, 255:red, 222; green, 222; blue, 222 }  ,fill opacity=1 ] (343.4,137.26) -- (352.97,130.4) .. controls (355.61,128.51) and (360.17,130.32) .. (363.16,134.45) .. controls (366.14,138.58) and (366.42,143.47) .. (363.78,145.36) -- (354.21,152.22) .. controls (351.57,154.11) and (347.01,152.3) .. (344.02,148.17) .. controls (341.04,144.04) and (340.76,139.15) .. (343.4,137.26) .. controls (346.04,135.37) and (350.61,137.18) .. (353.59,141.31) .. controls (356.58,145.44) and (356.85,150.32) .. (354.21,152.22) ;
\draw [color={rgb, 255:red, 208; green, 208; blue, 208 }  ,draw opacity=1 ][line width=2.25]    (360.21,136.97) -- (410.03,105.72) ;
\draw  [fill={rgb, 255:red, 255; green, 255; blue, 255 }  ,fill opacity=1 ] (403.4,105.68) .. controls (403.4,102.68) and (405.83,100.24) .. (408.84,100.24) .. controls (411.84,100.24) and (414.28,102.68) .. (414.28,105.68) .. controls (414.28,108.69) and (411.84,111.13) .. (408.84,111.13) .. controls (405.83,111.13) and (403.4,108.69) .. (403.4,105.68) -- cycle ; \draw   (403.4,105.68) -- (414.28,105.68) ; \draw   (408.84,100.24) -- (408.84,111.13) ;
\draw  [fill={rgb, 255:red, 0; green, 0; blue, 0 }  ,fill opacity=1 ] (414.28,105.84) .. controls (414.28,108.76) and (412.05,111.13) .. (409.29,111.13) -- (409.29,105.84) -- cycle ;
\draw  [fill={rgb, 255:red, 0; green, 0; blue, 0 }  ,fill opacity=1 ] (403.4,105.68) .. controls (403.4,105.68) and (403.4,105.68) .. (403.4,105.68) .. controls (403.4,105.68) and (403.4,105.68) .. (403.4,105.68) .. controls (403.39,102.76) and (405.7,100.39) .. (408.57,100.37) -- (408.59,105.66) -- cycle ;

\draw [color={rgb, 255:red, 224; green, 131; blue, 105 }  ,draw opacity=1 ][line width=0.75]    (395.38,74.29) -- (407.32,98.51) ;
\draw [shift={(394.05,71.6)}, rotate = 63.74] [fill={rgb, 255:red, 224; green, 131; blue, 105 }  ,fill opacity=1 ][line width=0.08]  [draw opacity=0] (3.57,-1.72) -- (0,0) -- (3.57,1.72) -- cycle    ;
\draw [color={rgb, 255:red, 169; green, 224; blue, 85 }  ,draw opacity=1 ][line width=0.75]    (409.95,29.24) -- (394.05,71.6) ;
\draw [shift={(411,26.43)}, rotate = 110.57] [fill={rgb, 255:red, 169; green, 224; blue, 85 }  ,fill opacity=1 ][line width=0.08]  [draw opacity=0] (3.57,-1.72) -- (0,0) -- (3.57,1.72) -- cycle    ;
\draw [color={rgb, 255:red, 96; green, 128; blue, 224 }  ,draw opacity=1 ][line width=1.5]    (413.52,24.93) -- (408.29,98.83) ;
\draw [shift={(413.8,20.94)}, rotate = 94.04] [fill={rgb, 255:red, 96; green, 128; blue, 224 }  ,fill opacity=1 ][line width=0.08]  [draw opacity=0] (4.64,-2.23) -- (0,0) -- (4.64,2.23) -- cycle    ;
\draw [color={rgb, 255:red, 169; green, 224; blue, 85 }  ,draw opacity=1 ][line width=0.75]    (462.55,62.63) -- (436.21,87.09) ;
\draw [shift={(462.55,62.63)}, rotate = 137.11] [color={rgb, 255:red, 169; green, 224; blue, 85 }  ,draw opacity=1 ][line width=0.75]    (7.06,-1.32) .. controls (5.47,-0.56) and (4.01,-0.12) .. (2.68,0) .. controls (4.01,0.12) and (5.47,0.56) .. (7.06,1.32)(4.37,-1.32) .. controls (2.78,-0.56) and (1.32,-0.12) .. (0,0) .. controls (1.32,0.12) and (2.78,0.56) .. (4.37,1.32)   ;
\draw [color={rgb, 255:red, 96; green, 128; blue, 224 }  ,draw opacity=1 ][line width=1.5]    (458.29,57.24) -- (414.77,100.64) ;
\draw [shift={(458.29,57.24)}, rotate = 135.07] [color={rgb, 255:red, 96; green, 128; blue, 224 }  ,draw opacity=1 ][line width=1.5]    (9.17,-1.71) .. controls (7.1,-0.72) and (5.21,-0.15) .. (3.49,0) .. controls (5.21,0.15) and (7.1,0.72) .. (9.17,1.71)(5.68,-1.71) .. controls (3.61,-0.72) and (1.72,-0.15) .. (0,0) .. controls (1.72,0.15) and (3.61,0.72) .. (5.68,1.71)   ;
\draw [color={rgb, 255:red, 224; green, 131; blue, 105 }  ,draw opacity=1 ][line width=0.75]    (436.21,87.09) -- (429.79,93.44) -- (419.72,103.4) ;
\draw [shift={(436.21,87.09)}, rotate = 135.32] [color={rgb, 255:red, 224; green, 131; blue, 105 }  ,draw opacity=1 ][line width=0.75]    (7.06,-1.32) .. controls (5.47,-0.56) and (4.01,-0.12) .. (2.68,0) .. controls (4.01,0.12) and (5.47,0.56) .. (7.06,1.32)(4.37,-1.32) .. controls (2.78,-0.56) and (1.32,-0.12) .. (0,0) .. controls (1.32,0.12) and (2.78,0.56) .. (4.37,1.32)   ;
\draw [color={rgb, 255:red, 108; green, 148; blue, 49 }  ,draw opacity=1 ][line width=0.75]    (306.76,181.91) -- (348,146.02) ;
\draw [shift={(348,146.02)}, rotate = 498.96] [color={rgb, 255:red, 108; green, 148; blue, 49 }  ,draw opacity=1 ][line width=0.75]    (7.06,-1.32) .. controls (5.47,-0.56) and (4.01,-0.12) .. (2.68,0) .. controls (4.01,0.12) and (5.47,0.56) .. (7.06,1.32)(4.37,-1.32) .. controls (2.78,-0.56) and (1.32,-0.12) .. (0,0) .. controls (1.32,0.12) and (2.78,0.56) .. (4.37,1.32)   ;
\draw [color={rgb, 255:red, 94; green, 94; blue, 94 }  ,draw opacity=1 ][line width=0.75]    (398.34,105.17) -- (312.43,110.15) ;
\draw [shift={(401.33,105)}, rotate = 176.69] [fill={rgb, 255:red, 94; green, 94; blue, 94 }  ,fill opacity=1 ][line width=0.08]  [draw opacity=0] (3.57,-1.72) -- (0,0) -- (3.57,1.72) -- cycle    ;
\draw [color={rgb, 255:red, 94; green, 94; blue, 94 }  ,draw opacity=1 ][line width=0.75]    (398.12,109.59) -- (356.06,135.93) ;
\draw [shift={(400.67,108)}, rotate = 147.95] [fill={rgb, 255:red, 94; green, 94; blue, 94 }  ,fill opacity=1 ][line width=0.08]  [draw opacity=0] (3.57,-1.72) -- (0,0) -- (3.57,1.72) -- cycle    ;
\draw [color={rgb, 255:red, 152; green, 152; blue, 152 }  ,draw opacity=1 ][line width=1.5]    (261.25,36.63) .. controls (278.75,25.88) and (285.75,43.63) .. (305.75,30.63) ;

\draw (238.32,119.07) node [anchor=north west][inner sep=0.75pt]   [align=left] {$\displaystyle \Vf{L_0}{\omega}{J_1}{B_1}$};
\draw (272.7,162.29) node [anchor=north west][inner sep=0.75pt]   [align=left] {$\displaystyle \Vf{L_0}{\omega}{J_2}{B_2}$};
\draw (414.88,4.36) node [anchor=north west][inner sep=0.75pt]   [align=left] {$\displaystyle \Vfdot{L_0}{p}{J_1}{B_2}$};
\draw (457.21,37.49) node [anchor=north west][inner sep=0.75pt]   [align=left] {$\displaystyle \Vf{L_0}{\omega}{J_1}{B_2}$};
\draw (338.99,82.06) node [anchor=north west][inner sep=0.75pt]   [align=left] {$\displaystyle \Vf{L_0}{p}{J_1}{B_2}$};
\draw (380.36,124.97) node [anchor=north west][inner sep=0.75pt]   [align=left] {$\displaystyle \Vf{L_0}{p}{J_2}{B_2}$};
\draw (415.22,109.89) node [anchor=north west][inner sep=0.75pt]   [align=left] {$\displaystyle \F{B_2}$};
\draw (355.51,153.89) node [anchor=north west][inner sep=0.75pt]   [align=left] {$\displaystyle \F{J_2}$};
\draw (278.97,91.17) node [anchor=north west][inner sep=0.75pt]   [align=left] {$\displaystyle \F{J_1}$};
\draw (280.37,18.75) node [anchor=north west][inner sep=0.75pt]   [align=left] {$\displaystyle \F{L_0}$};

\end{tikzpicture}
    \caption{Kinematics of a 2 DOF manipulator showing the angular and tangential velocity of the body frame $\F{B_2}$ due to the movement of the joints $\F{J_1}$ and $\F{J_2}$.}\label{fig:kinematics}
\end{figure}
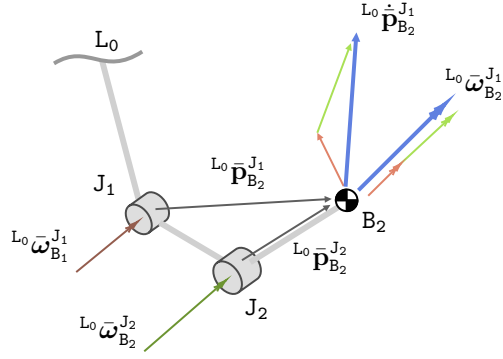

Let there be a manipulator with $\SET{J}=\left\{\F{J_1}, \F{J_2}, \dots, \F{J_{N}} \right\}$ joints, $\SET{L}=\left\{\F{L_0}, \F{L_1}, \dots, \F{L_{N}} \right\}$ links and $\SET{B}=\left\{\F{L_0}, \F{L_1}, \dots, \F{L_{N}} \right\}$ bodies forming a kinematic tree of nodes e.g. as shown in figure \ref{fig:transforms}. To traverse the kinematic tree, let there be a function $p(i)$ that retrieves an ordered list of all parents of node $i$ starting at node $i$ and ending with the base link $\F{L_0}$.

The transformation of each node $i$ in the tree with regard to $\F{L_0}$ is given by the product of each local transformation $\M{A}$ between the node $i$ and $\F{L_0}$
\begin{equation}
    \TR{A}{L_0}{i} = \prod_{\M{A} \in p(i)}{ \M{A} }.
    \label{eq:transf-parent}
\end{equation}

Generally, $\TR{A}{L_i}{B_i}$, $\TR{A}{L_i}{J_i}$ are constants defined by the physical configuration of the robot. On the other hand, $\TR{A}{J_i}{L_i}$ is a function of the joint position. For revolute joints, $\TR{A}{J_i}{L_i}\left(\theta_i\right)$ designates a rotation around the local joint axis ${}^\F{J_i}\V{n}_\F{J_i} \in \GROUP{R}^3, \NORM{\V{n}}=1$ (e.g.~the local z-axis), with an angle $\theta_i$, which corresponds to \eqref{eq:rodrigues} and can be encapsulated in the homogeneous transformation
\begin{equation}
    \TR{A}{J_i}{L_i}\left(\theta_i\right)= \begin{bmatrix}
        R\left(\theta_i,\Vf{J_i}{n}{}{J_i}\right) & \V{0} \\
        \V{0}                                     & 1
    \end{bmatrix}.
\end{equation}
Notice that although herein the focus is on revolute joints, a completely analogous procedure can be followed for prismatic joints.

The position of each node with respect to $\F{L_0}$ is obtained by the relation
\begin{equation}
    \begin{bmatrix} \Vf{L_0}{p}{L_0}{i} \\ 1 \end{bmatrix} = \TR{A}{L_0}{i} \begin{bmatrix}
        \V{0} \\ 1
    \end{bmatrix}.
\end{equation}

The angular velocity of $\F{B_i}$ relative to its parent joint $\F{J_i}$ and expressed in $\F{J_i}$ is denoted as $\Vf{J_i}{\omega}{J_i}{B_i}$ (fig.~\ref{fig:kinematics}), which can be expressed in terms of its rotation axis and angular velocity
\begin{equation}
    \Vf{J_i}{\omega}{J_i}{B_i} = \Vf{J_i}{n}{}{J_i} \dot{\theta}_\F{J_i}.
    \label{eq:angular_velocity_bi}
\end{equation}
It is clear that the body frame $\F{B_i}$, and the link $\F{L_i}$ must have the same angular velocity since $\TR{A}{B_i}{L_i}$ is constant, therefore \ $\Vf{J_i}{\omega}{J_i}{L_i} = \Vf{J_i}{\omega}{J_i}{B_i}$.

By multiplying with the rotational part $\TR{R}{L_0}{J_i}$ of $\TR{A}{L_0}{J_i}$, the angular velocity is obtained with respect to $\F{L_0}$:
\begin{equation}
    \Vf{L_0}{\omega}{J_i}{B_i} = \TR{R}{L_0}{J_i} \Vf{J_i}{\omega}{J_i}{B_i}.
    \label{eq:omega-identity}
\end{equation}
Using \eqref{eq:angular_velocity_bi}, it follows
\begin{equation}
    \Vf{L_0}{\omega}{J_i}{B_i} = \TR{R}{L_0}{J_i} \Vf{J_i}{n}{}{J_i} \dot{\theta}_\F{J_i}.
    \label{eq:sectional-angular-velocity}
\end{equation}
With all angular velocities expressed in the same frame of reference, the total angular velocity (relative to the base link) of each body is obtained by the sum of the angular velocities \eqref{eq:sectional-angular-velocity} over all of its ancestors:
\begin{equation}
    \Vf{L_0}{\omega}{L_0}{B_i} = \sum_{\F{J_j} \in p(B_i)}{\Vf{L_0}{n}{}{J_j} \dot{\theta}_\F{J_j}}.
    \label{eq:manipulator-angular-velocity}
\end{equation}
Rewriting \eqref{eq:manipulator-angular-velocity} as matrix-vector product then yields the Jacobian for rotation $\TR{J}{L_0}{\omega,i}$ for the $i$-th body in the chain. The $k$-th element of that matrix is given by
\begin{align}
    \TR{J}{L_0}{\omega,i,k} = \begin{cases}
        {}^\F{L_0}\V{n}_\F{J_k}, & \text{if } \F{J_k} \in p\left(B_i\right) \\
        \V{0}                    & \text{otherwise}
    \end{cases}.\label{eq:Jacobian-rotation}
\end{align}
Noncontributing joints are thus accounted by a $\V{0}$ contribution.

The linear, sectional velocity of the body $\Vfdot{L_0}{p}{J_i}{B_i}$ due to the rotational movement $\TR{A}{J_j}{B_i}$ of one of its ancestral joints $\F{J}_j \in p(\F{B}_i)$ is given by the tangential velocity
\begin{equation}
    \Vfdot{L_0}{p}{J_j}{B_i} = \Vf{L_0}{\omega}{J_j}{B_i} \times \underbrace{\left( \Vf{L_0}{p}{L_0}{J_j} - \Vf{L_0}{p}{L_0}{B_i} \right)}_{\Vf{L_0}{p}{J_j}{B_i}},
    \label{eq:sectional-velocity}
\end{equation}
where $\Vf{L_0}{p}{J_j}{B_i}$ defines the lever formed between the frames $\F{J_j}$ and $\F{B_i}$ (see figure~\ref{fig:kinematics}).
By using identity~\eqref{eq:omega-identity} and defining $\Vf{L_0}{r}{J_j}{B_i} = {}^\F{L_0}\V{n}_j \times \Vf{L_0}{p}{J_j}{B_i}$,~\eqref{eq:sectional-velocity} can be written as
\begin{equation}
    \Vfdot{L_0}{p}{J_j}{B_i} = \Vf{L_0}{r}{J_j}{B_i} \dot{\theta}_\F{J_j}.
    \label{eq:manipulator-linear-velocity}
\end{equation}
With all of the sectional velocities expressed with regard to $\F{L_0}$, adding them yields the total linear velocity of $\F{B_i}$
\begin{equation}
    \Vfdot{L_0}{p}{L_0}{B_i} = \sum_{\F{J_j} \in p(B_i)}{\Vf{\F{L_0}}{r}{J_j}{B_i} \dot{\theta}_\F{J_j}},
\end{equation}
This sum can also be written as $\Vfdot{L_0}{p}{L_0}{B_i} = \TR{J}{L_0}{t,i} \V{\theta}$, where $\TR{J}{L_0}{t,i}$ is the Jacobian matrix for translation. The $k$-th element of that matrix is given by
\begin{align}
    \TR{J}{L_0}{t,i,k} = \begin{cases}
        \Vf{L_0}{r}{J_k}{B_i}, & \text{if } \F{J_k} \in p\left(B_i\right) \\
        \V{0}                  & \text{otherwise}
    \end{cases}.\label{eq:Jacobian-translation}
\end{align}
Joints that are not an ancestor of $\F{B_i}$ do not add to its velocity, and their contribution is thus $\V{0}$.

Eq.~\eqref{eq:manipulator-linear-velocity} and~\eqref{eq:manipulator-angular-velocity} can thus be expressed in terms of the Jacobian matrices \eqref{eq:Jacobian-rotation}, \eqref{eq:Jacobian-translation}. The total linear velocity is given by
\begin{equation}
    \Vfdot{L_0}{p}{L_0}{B_i} = \TR{J}{L_0}{t,i} \dot{\V{\theta}},
    \label{eq:manipulator-linear-velocity-jacobian}
\end{equation}
with the total angular velocity $\Vf{L_0}{\omega}{}{B_i}$ of body $i$ being
\begin{equation}
    \Vf{L_0}{\omega}{L_0}{B_i} = \TR{J}{L_0}{\omega,i} \dot{\V{\theta}},
    \label{eq:manipulator-angular-velocity-jacobian}
\end{equation}
where $\V{\theta} = \begin{bmatrix}
        \theta_\F{J_1}, \ldots, \theta_\F{J_N}
    \end{bmatrix} \in \GROUP{R}^{N_J}$ is the vector of all joint positions.

\subsection{Kinematics of the Manipulator attached to the flying Base}
In the previous section, the kinematics of the manipulator was derived with regard to $\F{L_0}$. Attaching the manipulator to a floating base (the UAV here) endows the system with an additional 6 DOF.
To account for those 6 DOF, it suffices to describe the forward kinematics of the manipulator with regard to the inertial frame $\F{W}$, knowing that the floating base is located at a position $\V{p}$ from the origin and that its orientation with regard to $\F{W}$ is given by the unit quaternion $\Qx{q}$, essentially inserting a virtual 6-DOF joint in between $\F{W}$ and $\F{L_0}$ that is parametrized by $\V{p}$ and $\Qx{q}$.

Looking at the system from $\F{W}$, it is clear that the position $\Vf{W}{p}{L_0}{B_i}$ of each body of the manipulator in $\F{W}$ with regard to $\F{L_0}$ is superimposed by the position $\Vf{W}{p}{W}{L_0}$ of the floating base, resulting in the absolute link position
\begin{equation}
    \Vf{W}{p}{W}{B_i} = \Vf{W}{p}{W}{L_0} + \Vf{W}{p}{L_0}{B_i}.
\end{equation}
Each link's position in $\F{W}$ is obtained by rotating it by $\Qx{q}$, such as:
\begin{equation}
    \left(0,\Vf{W}{p}{L_0}{B_i}\right) = \Qx{q}{} \otimes \left(0,\Vf{L_0}{p}{L_0}{B_i}\right) \otimes \Qx{q}{}^*.\label{eq:link-position-inertial}
\end{equation}
The link velocity is then obtained from the time derivative of its position vector \eqref{eq:link-position-inertial}
\begin{equation}
    \Vfdot{W}{p}{W}{B_i} = \frac{d}{dt}\left(\Vf{W}{p}{W}{B_i}\right). \label{eq:link-linear-velocity-inertial}
\end{equation}
Inserting~\eqref{eq:link-position-inertial} in~\eqref{eq:link-linear-velocity-inertial} yields
\begin{equation}\begin{split}
        \left(0,\Vfdot{W}{p}{W}{B_i}\right)
        &= \left(0, \Vfdot{W}{p}{W}{L_0}\right) \\
        &+ \Qx{q} \otimes \left(0,\Vfdot{L_0}{p}{L_0}{B_i}\right) \otimes \Qx{q}^*  \\
        &+ \dot{\Qx{q}} \otimes \left(0,\Vf{L_0}{p}{L_0}{B_i}\right) \otimes \Qx{q}^* + \Qx{q} \otimes \left(0,\Vf{L_0}{p}{L_0}{B_i}\right) \otimes \dot{\Qx{q}}^*. \label{eq:link-velocity-inertial-quaternion}
    \end{split}\end{equation}
By multiplying the left resp.\ right side of the last two terms in~\eqref{eq:link-velocity-inertial-quaternion} by $\Qx{1}=\Qx{q}\otimes\Qx{q}^*$ the following expression is obtained:
\begin{equation}\begin{split}
        \left(0,\Vfdot{W}{p}{W}{B_i}\right)
        &= \left(0, \Vfdot{W}{p}{W}{L_0}\right) \\
        &+ \Qx{q} \otimes \left(0,\Vfdot{L_0}{p}{L_0}{B_i}\right) \otimes \Qx{q}^* \\
        &+ \Qx{q} \otimes \left[
            \Qx{q}^* \otimes \dot{\Qx{q}} \otimes \left(0,\Vf{L_0}{p}{L_0}{B_i}\right) \right.
            +\left.\left(0,\Vf{L_0}{p}{L_0}{B_i}\right) \otimes \dot{\Qx{q}}^* \otimes \Qx{q}
            \right] \otimes \Qx{q}^*.
    \end{split}
\end{equation}
Applying~\eqref{eq:wb} yields
\begin{equation}\begin{split}
        \left(0,\Vfdot{W}{p}{W}{B_i}\right)
        &= \left(0, \Vfdot{W}{p}{W}{L_0}\right) \\
        &+ \Qx{q} \otimes \left(0,\Vfdot{L_0}{p}{L_0}{B_i}\right) \otimes \Qx{q}^* \\
        &+ \Qx{q} \otimes \left[
            \left(0,\frac{1}{2} \Vf{L_0}{\omega}{L_0}{L_0}\right) \otimes \left(0,\Vf{L_0}{p}{L_0}{B_i}\right) \right.
            + \left. \left(0,\Vf{L_0}{p}{L_0}{B_i}\right) \otimes \left(0,-\frac{1}{2} \Vf{L_0}{\omega}{L_0}{L_0}\right)
            \right] \otimes \Qx{q}^*.
    \end{split}\end{equation}
Evaluating the two pure quaternion products using \eqref{eq:quaternion-product} then results in
\begin{equation}\begin{split}
        \left(0,\Vfdot{W}{p}{W}{B_i}\right)
        &= \left(0, \Vfdot{W}{p}{W}{L_0}\right) \\
        &+ \Qx{q} \otimes \left(0,\Vfdot{L_0}{p}{L_0}{B_i}\right) \otimes \Qx{q}^* \\
        &+ \Qx{q} \otimes \frac{1}{2} \left[
            \left(-\Vf{L_0}{\omega}{L_0}{L_0} \Vf{L_0}{p}{L_0}{B_i}, \Vf{L_0}{\omega}{L_0}{L_0} \times \Vf{L_0}{p}{L_0}{B_i}\right) \right.
            +\left. \left(\Vf{L_0}{\omega}{L_0}{L_0} \Vf{L_0}{p}{L_0}{B_i}, \Vf{L_0}{\omega}{L_0}{L_0} \times \Vf{L_0}{p}{L_0}{B_i}\right)
            \right] \otimes \Qx{q}^*,
    \end{split}\end{equation}
which further simplifies to
\begin{equation}\begin{split}
        \left(0,\Vfdot{W}{p}{W}{B_i}\right)
        &= \left(0, \Vfdot{W}{p}{W}{L_0}\right) \\
        &+ \Qx{q} \otimes \left(0,\Vfdot{L_0}{p}{L_0}{B_i}\right) \otimes \Qx{q}^* \\
        &+ \Qx{q} \otimes \left[
            \left(0, \Vf{L_0}{\omega}{L_0}{L_0} \times \Vf{L_0}{p}{L_0}{B_i}\right)
            \right] \otimes \Qx{q}^*.
    \end{split}\end{equation}
Using \eqref{eq:E}, \eqref{eq:G}, \eqref{eq:quaternion-rotmat} and \eqref{eq:cross-product}, it can further be transformed into
\begin{equation}\begin{split}
        \Vfdot{W}{p}{W}{B_i}
        &= \Vfdot{W}{p}{W}{L_0} \\
        &+ \TR{R}{W}{L_0} \Vfdot{L_0}{p}{L_0}{B_i} \\
        &- \TR{R}{W}{L_0} \CROSS{\Vf{L_0}{p}{L_0}{B_i}} \Vf{L_0}{\omega}{L_0}{L_0}.
    \end{split}\end{equation}
Introducing \eqref{eq:G} and \eqref{eq:manipulator-linear-velocity-jacobian} finally yields the complete equation of the linear velocity of $\F{B_i}$ in the inertial frame
\begin{gather}
    \begin{aligned}
        \Vfdot{W}{p}{W}{B_i} & = \Vfdot{W}{p}{W}{L_0}                                              & \text{(base translation)} \\
                             & + \TR{R}{W}{L_0} \TR{J}{L_0}{t,i} \dot{\V{\theta}}                  & \text{(joint rotation)}   \\
                             & - 2 \TR{R}{W}{L_0} \CROSS{\Vf{L_0}{p}{L_0}{B_i}} \M{G} \Qx{\dot{q}} & \text{(base rotation)}
        \label{eq:joint-velocity-translation-world}
    \end{aligned}
\end{gather}
showing the contributions of the moving base and the rotating links. In particular, the last term shows the component of the translational velocity due to the lever $\Vf{L_0}{p}{L_0}{B,i}$ and the angular velocity of the base.

The angular velocity of the bodies is simply the superposition of their own angular velocity and that of the base link in the inertial frame:
\begin{equation}
    \left(0,\Vf{W}{\omega}{W}{B_i}\right) = \left(0,\Vf{W}{\omega}{W}{L_0}\right) + \Qx{q} \otimes \left(0,\Vf{L_0}{\omega}{L_0}{B_i}\right) \otimes \Qx{q}^*.
\end{equation}
By inserting~\eqref{eq:body-angular-velocity-inertial},~\eqref{eq:manipulator-angular-velocity-jacobian} and writing the equation in matrix form then yields the expression for $\F{B_i}$'s angular velocity in world frame as the 4-vector
\begin{align}
    \begin{bmatrix}0 \\ \Vf{W}{\omega}{W}{B_i} \end{bmatrix} = 2\Q{q}{R}^\intercal \Qx{\dot{q}} + \begin{bmatrix}0 \\ \TR{R}{W}{L_0} \TR{J}{L_0}{\omega,i} \end{bmatrix} \dot{\V{\theta}}.
    \label{eq:joint-velocity-rotation-world}
\end{align}

\section{System Dynamics}
Using the kinematic equations of the system from the previous section, the system dynamics can now be obtained by leveraging the Lagrangian framework. In the following section, the direct dynamics are derived from the kinematic equations \eqref{eq:joint-velocity-translation-world} and \eqref{eq:joint-velocity-rotation-world}. Furthermore, the relation between the body- and generalized forces is established from the principle of virtual work. Lastly, those body forces are defined in the context of an AM.

\subsection{State Space and System Jacobian}
The state space $\V{z}$ of the combined system is defined by the choice of the (generalized) coordinates, namely the orientation given by the quaternion $\Qx{q}$, the position $\V{p}$ and the $N_J$ joint angles collected in $\V{\theta}$:
\begin{align}
    \V{z} = \begin{bmatrix}
        \V{x} & \Vfdot{}{x}{}{}
    \end{bmatrix},
    \label{eq:state-space}
\end{align}
with
\begin{align}
    \V{x} = \begin{bmatrix}
        \V{p} & \Qx{q} & \V{\theta}
    \end{bmatrix} \in \GROUP{R}^{7+N_J}. \label{eq:generalized-coordinates}
\end{align}

The rotational and translational velocity of each particle of the system can be written in form of $\V{v}=\TR{J}{W}{}\left(\V{x}\right)\dot{\V{x}}$, where $\V{v}$ is the vector of angular and linear velocities of each body of the system:
\begin{equation}
    \V{v} = {\left[\Vfdot{W}{p}{W}{B_0}, \dots, \Vfdot{W}{p}{W}{B_{N-1}}, \left(0, \Vfdot{W}{\omega}{W}{B_0}\right), \dots  \left(0, \Vfdot{W}{\omega}{W}{B_{N-1}}\right) \right]}.
\end{equation}
$\TR{J}{W}{}$ is the system's Jacobian matrix which is composed by the individual components of~\eqref{eq:joint-velocity-translation-world} and~\eqref{eq:joint-velocity-rotation-world}:
\begin{align}
    \TR{J}{W}{} = \begin{bmatrix}
        \TR{J}{W}{t,0}      \\ \vdots \\ \TR{J}{W}{t,N} \\
        \TR{J}{W}{\omega,0} \\ \vdots \\ \TR{J}{W}{t,N}
    \end{bmatrix}\in \GROUP{R}^{7N_B \times (7+N_J)},
\end{align}
with the individual Jacobian matrices for translation being
\begin{align}
    \TR{J}{W}{t,i} & = \begin{bmatrix}\begin{array}{c|c|c}
            \M{1}_{3} & -2 \TR{R}{W}{L_0} \CROSS{\Vf{L_0}{p}{L_0}{B_i}} \M{G} & \TR{R}{W}{L_0}\TR{J}{L_0}{t,i}
        \end{array}\end{bmatrix} \in \GROUP{R}^{3 \times (7+N_J)},
\end{align}
and for rotation being
\begin{align}
    \TR{J}{W}{\omega,i} & = \begin{bmatrix}\begin{array}{c|c|c}
            \M{0}_{4\times3} & 2 \Q{q}{R}^\intercal & \begin{bmatrix} 0 \\ \TR{R}{W}{L_0}\TR{J}{L_0}{\omega,i} \end{bmatrix}
        \end{array}\end{bmatrix} \in \GROUP{R}^{4 \times (7+N_J)}.
\end{align}

\subsection{Equations of Motion}
Solving the Lagrangian equations is an algorithmically simple way of retrieving the dynamics equations. It does, however, come with some requirements concerning the choice of the coordinates. More precisely, the coordinates need to be independent and complete. It is clear that the chosen coordinates \eqref{eq:generalized-coordinates} are complete and can represent any configuration of the system. The requirement of independent coordinates is however not met e.g.\ fixing the coordinates $q_x$, $q_y$, $q_z$ of $\Qx{q}$ will also fully constrain $q_w$ by the unity constraint $\NORM{\Qx{q}}=1$. For the time being, it will be assumed that all the coordinates are independent. The holonomic constraint will be dealt with at a later stage.

In the Lagrangian framework, the system dynamics are obtained by solving the Euler-Lagrange equations
\begin{equation}
    \frac{d}{dt}\frac{\partial \mathcal{L}}{\partial \dot{\V{x}}} - \frac{\partial \mathcal{L}}{\partial \V{x}} = \V{f}_x + \V{f}_c, \label{eq:lagrange}
\end{equation}
where $\mathcal{L}=\mathsf{E_{kin}}-\mathsf{E_{pot}}$ is the Lagrangian, $\V{x}$ the state space of the system, $\V{f}_x \in \GROUP{R}^{N_x}$ are the generalized forces resp. torques along the generalized coordinates, and $\V{f}_c \in \GROUP{R}^{N_x}$ mark the constraint forces, pulling the system towards the manifold defined by the holonomic constraints. Herein, the constraint forces arise by the virtue of the unit quaternion coordinates not being independent.

In the case of mechanical systems, the solution of \eqref{eq:lagrange} takes the following particular form~\cite{BrunoSiciliano2008}:
\begin{align}
    \M{M}\left(\V{x}\right) \ddot{\V{x}} + \M{C}\left(\V{x}, \dot{\V{x}}\right) \dot{\V{x}} + \V{g}\left(\V{x}\right) = \V{f}_{x} + \V{f}_{c},
    \label{eq:system-dynamics}
\end{align}
where $\M{M}\in \GROUP{R}^{N_x \times N_x}$ is the mass matrix, $\M{C}\in \GROUP{R}^{N_x \times N_x}$ the Coriolis matrix and $\V{g}\in \GROUP{R}^{N_x}$ the vector of gravitational terms. The total kinetic energy of the system is given by
\begin{align}
    \mathsf{E_{kin}} = \frac{1}{2} \dot{\V{x}} \TR{J}{W}{}^\intercal \M{M}_L \TR{J}{W}{} \dot{\V{x}} \label{eq:E-kin}
\end{align}
with $\M{M}_L$ being the generalized inertia matrix of the system represented by the block diagonal formed by the individual body masses $m_i$ and the moments of inertia ${}^\F{W}\M{\Theta}_i$ in the world frame \cite{Moller2012, Udwadia2016}. It is defined as
\begin{equation}\begin{split}
        \M{M}_L =   \text{blockdiag}\left( \M{1}_{3} m_0, \dots, \M{1}_{3} m_{N_B-1},
        {}^\F{W}\M{\Theta}_0, \dots, {}^\F{W}\M{\Theta}_{N_B-1} \right),
    \end{split}\end{equation}
with ${}^\F{W}\M{\Theta}_i$ being linked to the classical inertia tensor ${}^\F{W}\M{\Phi}_i$ by
\begin{align}
    {}^\F{W}\M{\Theta}_i = \begin{bmatrix}
        \nu   & \V{0}              \\
        \V{0} & {}^\F{W}\M{\Phi}_i
    \end{bmatrix} \in \GROUP{R}^{4\times4}.
\end{align}
Therein, $\nu$ is associated with the scaling DOF of a body, such would be the case without enforcing $\NORM{\Qx{q}}=1$. However, since $\Qx{q}$ is assumed to be of unit length, the bodies of the system are rigid and $\nu$ does not come into play. As such, $\nu$ can be defined as any positive number.

Solving \eqref{eq:lagrange} directly via calculating $\mathcal{L}$ becomes impractical resp. computationally expensive and thus time-consuming for larger systems. A much faster approach is the factorization into the individual system matrices \eqref{eq:system-dynamics}, which has the additional benefit of not having to separate the individual components into their respective matrix. Once the expression for the system's mass matrix is obtained, the Coriolis matrix can be obtained from its Christoffel symbols. The gravitational terms are obtained from the potential energy, which is usually a relatively small expression and thus inexpensive to compute.

By analyzing \eqref{eq:system-dynamics}, \eqref{eq:E-kin}, it can be seen that the only resulting term in $\ddot{\V{x}}$, corresponding to the system's mass matrix $\M{M}$ (symmetric, positive semi-definite \cite{Schutte2011}), is obtained from
\begin{align}
    \M{M} = \TR{J}{W}{}^\intercal \M{M}_L \TR{J}{W}{},
    \label{eq:system-mass-matrix}
\end{align}

The system's Coriolis matrix $\M{C}\in\GROUP{R}^{N_x\times N_x}$ containing the centrifugal and centripetal terms is obtained by calculating the Christoffel symbols first kind from the mass matrix~\eqref{eq:system-mass-matrix}~\cite{BrunoSiciliano2008}. Its individual components are given by
\begin{align}
    \M{C}_{ij} =
    \frac{1}{2} \sum_{k=1}^{N_x} \left( \frac{\partial\M{M}_{ij}}{\partial \V{x}_{k}} + \frac{\partial\M{M}_{ik}}{\partial \V{x}_{j}} - \frac{\partial\M{M}_{jk}}{\partial \V{x}_{i}} \right) \dot{\V{x}}_k.
    \label{eq:system-coriolis-matrix}
\end{align}
The definition of the Coriolis matrix is not unique, and different formulations are possible (e.g.~\cite{Bjerkeng2012}) having different properties. The choice herein has the property of $\dot{\M{M}}-2\M{C}$ being skew-symmetric, a useful property for controller design~\cite{BrunoSiciliano2008}.

Lastly, the system's gravity vector $\V{g}$ is obtained from the gradient of the potential energy
\begin{align}
    \mathsf{E_{pot}}\left(\V{x}\right) = \sum_{i \in N}{m_i \V{g}_0 \Vf{W}{p}{W}{B_i}},
\end{align}
with $\V{g}_0$ being the gravitational acceleration. The vector $\V{g}$ contains the gravitational terms and thus accounts for the shifting center of gravity due to the motion of the manipulator. It is obtained from the gradient of the potential energy by calculating
\begin{align}
    \V{g} =  \nabla \mathsf{E_{pot}} = \frac{\partial\mathsf{E_{pot}}}{\partial \V{x}}.
\end{align}

An alternative to those equations is presented in \cite{Wang2019}, where the system's mass matrix $\M{M}$ and lumped Coriolis vector $\V{h}$ are obtained from the system's total kinetic energy by calculating
\begin{align}
    \M{M} & = \frac{\partial}{\partial \dot{\V{x}}}\left( \frac{\partial \mathsf{E_{kin}}}{\partial \dot{\V{x}}} \right), \label{eq:mass-matrix-wang2019}                                                              \\
    \V{h} & = \frac{\partial}{\partial \V{x}}\left( \frac{\partial \mathsf{E_{kin}}}{\partial \dot{\V{x}}} \right) \dot{\V{x}} - \frac{\partial \mathsf{E_{kin}}}{\partial \V{x}}. \label{eq:coriolis-vector-wang2019}
\end{align}
The vector $\V{h}$ is equal to $\M{C}\dot{\V{x}}$.

\subsection{Force Mapping}
The term $\V{f}_x$ in~\eqref{eq:system-dynamics} is given in terms of forces along the generalized coordinates. In many situations, it is very convenient to have a mapping from local body forces to generalized forces. This mapping will greatly facilitate the application of the forces and torques originating from the propulsion system of the AM.

By the choice of the generalized coordinates, the torques applied to the AM have to be specified along the four coordinates of the quaternion. The principle of virtual work is used to establish a relationship between the quaternion torques and the Cartesian torques along the body-fixed axis.

The principle of virtual work states that the work $\delta W$ is equal to the force $\V{f}$ acting on a body along its virtual displacement $\delta\V{s}$:
\begin{align}
    \delta W = \V{f} \cdot \delta \V{s}.
\end{align}
In a completely analogous manner, the work performed by the torque $\Vf{L_0}{f}{}{q}$ acting on the rotation $\delta\V{\varphi}=\V{n}\delta\varphi$ is defined as
\begin{align}
    \delta W_q = \Vf{L_0}{f}{}{q} \cdot \V{n} \delta\varphi.
    \label{eq:virtual-work-rotation}
\end{align}
Following the approach in~\cite{Graf2008} stating that for small changes in rotations (i.e., $\NORM{\Qx{q}_\delta}=1$), the following approximation holds:
\begin{align}
    \Qx{q} + \delta \Qx{q} \approx \Qx{q} \otimes \Qx{q}_\delta,
    \label{eq:small_variation}
\end{align}
where $\delta \Qx{q}$ is a small variation and $\Qx{q}_\delta$ is a small change of rotation from $\Qx{q}$.
Left multiplying both sides of that relation with $\Qx{q}^*$ yields:
\begin{align}
    \left(1,\V{0}\right) + \Qx{q}^* \otimes \delta \Qx{q} \approx \Qx{q}_\delta.
    \label{eq:preliminary-result-work}
\end{align}
Using the axis angle relation \eqref{eq:axis-angle-notation}, $\Qx{q}_\delta$, under consideration of small angles, can be approximated as
\begin{align}
    \Qx{q}_\delta \approx \left(1, \frac{1}{2}\V{n}\delta\varphi\right). \label{eq:small-angle-approx}
\end{align}
Thus, \eqref{eq:small-angle-approx} in \eqref{eq:preliminary-result-work} results in
\begin{align}
    2 \left(\Qx{q}^* \otimes \delta \Qx{q}\right) \approx \left(0, \V{n}\delta\varphi\right).
    \label{eq:preliminary-result-work2}
\end{align}
Writing \eqref{eq:virtual-work-rotation} as the dot product of two pure quaternions
\begin{align}
    \delta W_q \approx \left(0, \Vf{L_0}{f}{}{q}\right) \cdot \left(0, \V{n}\delta\varphi\right),
\end{align}
and by inserting \eqref{eq:preliminary-result-work2} the relation
\begin{align}
    \delta W_q \approx \left(0, \Vf{L_0}{f}{}{q}\right) \cdot 2 \left(\Qx{q}^* \otimes \delta \Qx{q}\right)
\end{align}
is obtained. Using \eqref{eq:G} yields
\begin{align}
    \delta W_q \approx 2 \Vf{L_0}{f}{}{q} \cdot \M{G} \delta \Qx{q}. \label{eq:work1}
\end{align}
With the help of the dot product property $\V{x} \cdot \M{A}\V{y}=\M{A}^\intercal \V{x}\cdot\V{y}$, \eqref{eq:work1} becomes
\begin{align}
    \delta W_q \approx 2 \M{G}^\intercal \Vf{L_0}{f}{}{q} \cdot \delta \Qx{q}, \label{eq:quaternion-force-mapping}
\end{align}
wherein $2\M{G}^\intercal$ can be identified as the mapping between Cartesian torques in body frame and the generalized quaternion torques.
The forces along the body's fixed axis can be mapped to the generalized forces with
\begin{align}
    \Vf{W}{f}{}{xyz} = \TR{R}{W}{L_0} \Vf{L_0}{f}{}{xyz}\label{eq:body-forces-to-generalized}.
\end{align}
The mapping of joint torques to generalized joint forces is trivial:
\begin{align}
    \Vf{W}{f}{}{J} = \M{1}_{N_J} \Vf{L_0}{f}{}{J}\label{eq:joint-forces-to-generalized}.
\end{align}

The results \eqref{eq:quaternion-force-mapping}, \eqref{eq:body-forces-to-generalized} and \eqref{eq:joint-forces-to-generalized} are summarized in the following block diagonal force mapping matrix $\M{M}_F$, mapping from body forces to generalized forces $\V{f}_x$:
\begin{align}
    \M{M}_F =
    \text{blockdiag}\left(\TR{R}{W}{L_0}, 2\M{G}^\intercal, \M{1}_{N_J}\right),
    \label{eq:force-mapping-matrix}
\end{align}
such that the generalized forces in~\eqref{eq:system-dynamics} can be written as
\begin{align}
    \V{f}_x = \M{M}_F \V{f}_B,
\end{align}
with $\Vf{L_0}{f}{}{B}$ being the body forces composed by the forces $\Vf{L_0}{f}{}{xyz}$ and torques $\Vf{L_0}{f}{}{q}$ in resp.~about the local $x$, $y$ and $z$ body axis and the joint torques $\Vf{L_0}{f}{}{J}$. Those forces typically originate from the propulsion system but can also include aerodynamic effects if so desired. The body force vector is thus defined by
\begin{align}
    \Vf{L_0}{f}{}{B} = \begin{bmatrix}
        \Vf{L_0}{f}{}{xyz} & \Vf{L_0}{f}{}{q} & \Vf{L_0}{f}{}{J}
    \end{bmatrix}. \label{eq:body-forces}
\end{align}

\subsection{Propulsion Forces}\label{sec:body-forces}
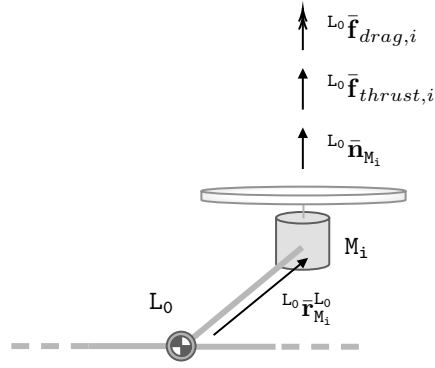
\begin{figure}
    \centering
  
\tikzset {_rcne72zbr/.code = {\pgfsetadditionalshadetransform{ \pgftransformshift{\pgfpoint{0 bp } { 0 bp }  }  \pgftransformscale{1 }  }}}
\pgfdeclareradialshading{_df1yzevas}{\pgfpoint{0bp}{0bp}}{rgb(0bp)=(1,1,1);
rgb(17.05357142857143bp)=(1,1,1);
rgb(25bp)=(0.93,0.93,0.93);
rgb(400bp)=(0.93,0.93,0.93)}
\tikzset{every picture/.style={line width=0.75pt}} 

\begin{tikzpicture}[x=0.75pt,y=0.75pt,yscale=-1,xscale=1]

\draw [color={rgb, 255:red, 184; green, 184; blue, 184 }  ,draw opacity=1 ][line width=2.25]    (396.24,190.7) -- (349.62,190.7) ;
\draw [color={rgb, 255:red, 184; green, 184; blue, 184 }  ,draw opacity=1 ][line width=2.25]    (348.63,190.38) -- (302,190.38) ;
\draw  [color={rgb, 255:red, 93; green, 93; blue, 93 }  ,draw opacity=1 ][fill={rgb, 255:red, 225; green, 225; blue, 225 }  ,fill opacity=1 ] (423.33,125.6) -- (423.33,148.93) .. controls (423.33,150.54) and (417.33,151.85) .. (409.92,151.85) .. controls (402.51,151.85) and (396.5,150.54) .. (396.5,148.93) -- (396.5,125.6) .. controls (396.5,123.99) and (402.51,122.68) .. (409.92,122.68) .. controls (417.33,122.68) and (423.33,123.99) .. (423.33,125.6) .. controls (423.33,127.21) and (417.33,128.52) .. (409.92,128.52) .. controls (402.51,128.52) and (396.5,127.21) .. (396.5,125.6) ;
\draw [color={rgb, 255:red, 179; green, 179; blue, 179 }  ,draw opacity=1 ]   (410.33,112.68) -- (410.33,125.85) ;
\draw [color={rgb, 255:red, 184; green, 184; blue, 184 }  ,draw opacity=1 ][line width=2.25]    (409.67,140.52) -- (350,190) ;
\draw  [color={rgb, 255:red, 93; green, 93; blue, 93 }  ,draw opacity=1 ][fill={rgb, 255:red, 178; green, 178; blue, 178 }  ,fill opacity=1 ] (341.35,190.38) .. controls (341.35,186.36) and (344.61,183.1) .. (348.63,183.1) .. controls (352.64,183.1) and (355.9,186.36) .. (355.9,190.38) .. controls (355.9,194.39) and (352.64,197.65) .. (348.63,197.65) .. controls (344.61,197.65) and (341.35,194.39) .. (341.35,190.38) -- cycle ;
\path  [shading=_df1yzevas,_rcne72zbr] (461.83,112.12) -- (461.83,116.58) .. controls (461.83,117.81) and (438.78,118.81) .. (410.33,118.81) .. controls (381.89,118.81) and (358.83,117.81) .. (358.83,116.58) -- (358.83,112.12) .. controls (358.83,110.89) and (381.89,109.89) .. (410.33,109.89) .. controls (438.78,109.89) and (461.83,110.89) .. (461.83,112.12) .. controls (461.83,113.35) and (438.78,114.35) .. (410.33,114.35) .. controls (381.89,114.35) and (358.83,113.35) .. (358.83,112.12) ; 
 \draw  [color={rgb, 255:red, 145; green, 145; blue, 145 }  ,draw opacity=1 ] (461.83,112.12) -- (461.83,116.58) .. controls (461.83,117.81) and (438.78,118.81) .. (410.33,118.81) .. controls (381.89,118.81) and (358.83,117.81) .. (358.83,116.58) -- (358.83,112.12) .. controls (358.83,110.89) and (381.89,109.89) .. (410.33,109.89) .. controls (438.78,109.89) and (461.83,110.89) .. (461.83,112.12) .. controls (461.83,113.35) and (438.78,114.35) .. (410.33,114.35) .. controls (381.89,114.35) and (358.83,113.35) .. (358.83,112.12) ; 

\draw    (365,184.83) -- (410.34,147.73) ;
\draw [shift={(412.67,145.83)}, rotate = 500.71] [fill={rgb, 255:red, 0; green, 0; blue, 0 }  ][line width=0.08]  [draw opacity=0] (5.36,-2.57) -- (0,0) -- (5.36,2.57) -- cycle    ;
\draw    (410,101) -- (410,83) ;
\draw [shift={(410,80)}, rotate = 450] [fill={rgb, 255:red, 0; green, 0; blue, 0 }  ][line width=0.08]  [draw opacity=0] (5.36,-2.57) -- (0,0) -- (5.36,2.57) -- cycle    ;
\draw [color={rgb, 255:red, 184; green, 184; blue, 184 }  ,draw opacity=1 ][line width=2.25]  [dash pattern={on 6.75pt off 4.5pt}]  (302,190.38) -- (260,190.38) ;
\draw [color={rgb, 255:red, 184; green, 184; blue, 184 }  ,draw opacity=1 ][line width=2.25]  [dash pattern={on 6.75pt off 4.5pt}]  (438.24,190.7) -- (396.24,190.7) ;
\draw [color={rgb, 255:red, 0; green, 0; blue, 0 }  ,draw opacity=1 ]   (410,71) -- (410,53) ;
\draw [shift={(410,50)}, rotate = 450] [fill={rgb, 255:red, 0; green, 0; blue, 0 }  ,fill opacity=1 ][line width=0.08]  [draw opacity=0] (5.36,-2.57) -- (0,0) -- (5.36,2.57) -- cycle    ;
\draw [color={rgb, 255:red, 0; green, 0; blue, 0 }  ,draw opacity=1 ]   (410,41) -- (410,20) ;
\draw [shift={(410,20)}, rotate = 450] [color={rgb, 255:red, 0; green, 0; blue, 0 }  ,draw opacity=1 ][line width=0.75]    (10.58,-1.97) .. controls (8.19,-0.84) and (6.01,-0.18) .. (4.03,0) .. controls (6.01,0.18) and (8.19,0.84) .. (10.58,1.97)(6.56,-1.97) .. controls (4.17,-0.84) and (1.99,-0.18) .. (0,0) .. controls (1.99,0.18) and (4.17,0.84) .. (6.56,1.97)   ;
\draw  [color={rgb, 255:red, 93; green, 93; blue, 93 }  ,draw opacity=1 ][fill={rgb, 255:red, 255; green, 255; blue, 255 }  ,fill opacity=1 ] (343.65,190.38) .. controls (343.65,187.62) and (345.88,185.39) .. (348.64,185.39) .. controls (351.39,185.39) and (353.62,187.62) .. (353.62,190.38) .. controls (353.62,193.13) and (351.39,195.36) .. (348.64,195.36) .. controls (345.88,195.36) and (343.65,193.13) .. (343.65,190.38) -- cycle ; \draw  [color={rgb, 255:red, 93; green, 93; blue, 93 }  ,draw opacity=1 ] (343.65,190.38) -- (353.62,190.38) ; \draw  [color={rgb, 255:red, 93; green, 93; blue, 93 }  ,draw opacity=1 ] (348.64,185.39) -- (348.64,195.36) ;
\draw  [color={rgb, 255:red, 93; green, 93; blue, 93 }  ,draw opacity=1 ][fill={rgb, 255:red, 93; green, 93; blue, 93 }  ,fill opacity=1 ] (353.63,190.51) .. controls (353.63,190.51) and (353.63,190.51) .. (353.63,190.51) .. controls (353.63,193.19) and (351.58,195.36) .. (349.05,195.36) -- (349.05,190.51) -- cycle ;
\draw  [color={rgb, 255:red, 93; green, 93; blue, 93 }  ,draw opacity=1 ][fill={rgb, 255:red, 93; green, 93; blue, 93 }  ,fill opacity=1 ] (343.65,190.37) .. controls (343.65,190.37) and (343.65,190.37) .. (343.65,190.37) .. controls (343.63,187.7) and (345.76,185.51) .. (348.39,185.5) -- (348.41,190.36) -- cycle ;

\draw (331,162) node [anchor=north west][inner sep=0.75pt]   [align=left] {$\displaystyle \F{L_0}$};
\draw (429.67,134.35) node [anchor=north west][inner sep=0.75pt]   [align=left] {$\displaystyle \F{M_i}$};
\draw (421,82) node [anchor=north west][inner sep=0.75pt]   [align=left] {$\displaystyle \Vf{L_0}{n}{}{M_i}$};
\draw (398,162) node [anchor=north west][inner sep=0.75pt]   [align=left] {$\displaystyle \Vf{L_0}{r}{L_0}{M_i}$};
\draw (421,52) node [anchor=north west][inner sep=0.75pt]   [align=left] {$\displaystyle {}^\F{L_0}\V{f}_{thrust,i}$};
\draw (421,22) node [anchor=north west][inner sep=0.75pt]   [align=left] {$\displaystyle {}^\F{L_0}\V{f}_{drag,i}$};

\end{tikzpicture}
    \caption{The propulsion system. The thrust force along the motor-spin normal and the associated drag of the propeller generate a net force acting on $\F{L_0}$.}\label{fig:motor_forces}
\end{figure}

The body forces \eqref{eq:body-forces} are defined as a function of the thrust of the AM's propulsion system. Given a UAV with $N_M$ propulsion devices, then the forces from the propulsion system acting on the body of the AM are obtained from the following relations
\begin{subequations}
    \begin{align}
        \Vf{L_0}{f}{}{xyz} & = \sum_i^{N_M} {}^\F{L_0}\V{f}_{thrust,i}                                                                                                                  \\
        \Vf{L_0}{f}{}{q}   & = \sum_i^{N_M} {}^\F{L_0}\V{f}_{rot,i} = {\sum_i^{N_M} \left( \Vf{L_0}{r}{L_0}{M,i} \times {}^\F{L_0}\V{f}_{thrust,i} + {}^\F{L_0}\V{f}_{drag,i} \right)}.
    \end{align}
\end{subequations}
Therein $\V{f}_{thrust,i}$ designates the thrust force vector and $\V{f}_{drag,i}$ the drag vector (reaction torque). The vector $\Vf{L_0}{r}{L_0}{M,i}$ designates the lever formed between the UAV's body and the motor frame (see figure \ref{fig:motor_forces}). The mapping matrix $\M{M}_{mot}$, also commonly referred to as efficiency resp. allocation matrix (used by the mixer in many controllers) contains the contribution of each motor to each force resp. torque along each body axis. It is constant (unless the motors themselves can tilt) and defined by
\begin{equation}
    \M{M}_{mot} = \text{blockdiag}\left(
    \begin{bmatrix}
        {}^\F{L_0}\V{f}_{thrust,1}, \ldots, {}^\F{L_0}\V{f}_{thrust,N_M}
    \end{bmatrix},
    \begin{bmatrix}
        {}^\F{L_0}\V{f}_{rot,1}, \ldots, {}^\F{L_0}\V{f}_{rot,N_M}
    \end{bmatrix},
    \M{1}_N
    \right)\label{eq:body-force-to-motor-mapping}
\end{equation}
and can be used to map body forces to motor forces.

Both quantities are a function of the angular velocity $\omega_\F{M_i}$ of the propeller \cite{Kotarski2017} as in (see figure~\ref{fig:problem-illutration})
\begin{subequations}
    \begin{align}
        {}^\F{L_0}\V{f}_{thrust,i} & = \Vf{L_0}{n}{}{M_i} k_t \omega_\F{M_i}^2, \label{eq:prop_thrust}         \\
        {}^\F{L_0}\V{f}_{drag,i}   & = \Vf{L_0}{n}{}{M_i} k_p s_\F{M_i} \omega_\F{M_i}^2, \label{eq:prop_drag}
    \end{align}  \label{eq:motor-thrust-rpm}
\end{subequations}
where $\Vf{L_0}{n}{}{M,i}$ is the motor's normalized axis of rotation. The constants $k_t$ and $k_p$ for thrust and drag can be identified experimentally.
Furthermore, since $\omega_\F{M_i}$ is rarely known and only very few ESCs provide telemetry resp.~accept inputs as RPM references,~\eqref{eq:prop_thrust},~\eqref{eq:prop_drag}, can also be written as a function of the flight controller's output PWM signal $u_{M,i}=\left[0,1\right]$, which yields the roughly linear relationship~\cite{Kotarski2017},\cite{Yoon2015}:
\begin{subequations}
    \begin{align}
        {}^\F{L_0}\V{f}_{thrust,i} & \approx \Vf{L_0}{n}{}{M_i} \hat{k}_t u_\F{M_i},           \\
        {}^\F{L_0}\V{f}_{drag,i}   & \approx \Vf{L_0}{n}{}{M_i} \hat{k}_p s_\F{M_i} u_\F{M_i},
    \end{align}
\end{subequations}
where $s_{M,i} = \left\{1,-1\right\}$, for CW resp. CCW, designates the spin direction of the motor. Notice that for tilting propeller configurations, the normal vector $\Vf{L_0}{n}{}{M_i}$ would be a function of the tilt angles commanded to the tilting mechanism.

A real motor cannot instantaneously change its velocity; therefore, it is common practice to model the motor dynamics as a first-order system \cite{Yoon2015}:
\begin{equation}
    G\left(s\right) = \frac{K}{1+Ts}, \label{eq:motor-system}
\end{equation}
where $K$ is the maximum RPM and $T$ is the time constant of the motor-propeller system.

Robotic joints are generally more complex and have a non-negligible amount of friction and damping due to their internal mechanics. Their modeling highly depends on their internals and thus has to be seen on a case-by-case basis. For the sake of simplicity, the model in \eqref{eq:motor-system} is also used for the robotic joints throughout this work.

\section{Simulation \& Constraints}
The numeric simulation of the AM is a very helpful or even critical step before deploying an AM in practice. Furthermore, there are applications in deep reinforcement learning where the model needs to be simulated such that the controller can be learned from the model dynamics \cite{Manukyan2019}. The required equations for numerical simulation, respecting the holonomic constraints, are derived herein.

The system dynamics equation \eqref{eq:system-dynamics} can be written in form of the differential algebraic equation (DAE) \cite{Tassora2001}
\begin{numcases}{}
    \M{M} \ddot{\V{x}} + \M{C} \dot{\V{x}} + \V{g} - \V{q}_{F} - \lambda\M{J}_{\V{\phi}} = \V{0}, \\
    \V{\phi} = \V{0},
\end{numcases}
where $\V{\phi}$ are the holonomic constraints imposed on the system, $\M{J}_{\V{\phi}}$ designates the Jacobian matrix of $\V{\phi}$ and $\lambda$ is a Lagrange multiplier. The constraint vector contains (at least) the unity constraint originating from the unit quaternion:
\begin{align}
    \V{\phi}\left(\V{x}\right) = \left[\NORM{\Qx{q}} - 1\right].
\end{align}

The system equation can be used for numerical simulation by solving~\eqref{eq:system-dynamics} for $\ddot{\V{x}}$ and implementing it as first-order ODEs:
\begin{numcases}{\dot{\V{z}}=}
    \dot{\V{x}} = \V{v},\label{eq:solution-velocity}
    \\
    \dot{\V{v}} = \V{a},\label{eq:solution-accel}
\end{numcases}
wherein $\V{a}=\ddot{\V{x}}$ marks the unconstrained acceleration:
\begin{equation}
    \V{a} = \M{M}^{-1} \left[\V{f}_x - \M{C} \dot{\V{x}} - \V{g} \right]. \label{eq:forw-dynamics}
\end{equation}
Naturally, such a system would not be constrained to the manifold defined by the constraint $\V{\phi}=0$, violating the unit length assumption of the quaternion, thus degrading the physical fidelity of the system up to a point where it is completely disjoint from reality. It is, therefore, mandatory to constrain the system such that the unit length assumption holds. Generally, this can be seen as a holonomic constraint problem, which is typically addressed by Lagrangian multipliers within the Lagrangian framework.

Herein, we opted for a general-purpose solution to this holonomic constraints problem as discussed in \cite{Baumgarte1972},\cite{Yoon1992} and \cite{Braun2009}, in the spirit that it provides additional value, e.g., in a reinforcement learning context, where it might be useful to temporarily stiffen the joints to speed up the training process. The formulation is, however, fairly heavy and the authors of \cite{Tassora2001, Sherif2015} specifically addressed the problem of the quaternion unit constraint and provided a numerically faster method. The general idea behind those methods is, however, the same and consists of performing small corrective actions orthogonal to the manifolds defined by the constraints for both the system's velocities and positions. The authors of \cite{Moller2012} showed how to calculate the lagrangian multipliers directly. However, this requires splitting the system into translational and rotational parts, which is unfeasible here due to the coupled nature of the system.

According to \cite{Braun2009}, the holonomic constraint is enforced onto the system by constraining the accelerations and velocities, which are obtained by the sum of the unconstrained acceleration $\V{a}$ resp. velocity $\V{v}$ and a corrective term. The system respecting the holonomic constraints is thus given by
\begin{subnumcases}{\dot{\V{z}}_{\V{\phi}}=\label{eq:constrained-system}}
    \dot{\V{x}} = \V{v} + \M{M}^{-1/2} \M{B}^+ \left(\V{b}_{q} - \M{A} \dot{\V{v}} - \V{\phi} / \mathbf{dt} \right), \label{eq:solution-velocity-constrained} \\
    \dot{\V{v}} = \V{a} + \M{M}^{-1/2} \M{B}^+ \left(\V{b}_{v} - \M{A} \V{a} - \dot{\V{\phi}} / \mathbf{dt} \right),\label{eq:solution-acceleration-constrained}
\end{subnumcases}
where ${\left(.\right)}^+$ designates the Moore-Penrose ´pseudo´ inverse. The remaining terms are computed as
\begin{subequations} \label{eq:baumgarte-constraint-equations}
    \begin{align}
        \M{A}     & = \frac{\partial \V{\phi}}{\partial \V{x}} \in \GROUP{R}^{N_\phi \times N_x},                                                                                                                                              \\
        \M{B}     & = \M{A} \M{M}^{-1/2} \in \GROUP{R}^{N_\phi \times N_x},                                                                                                                                                                    \\
        \V{b}_{q} & = -\frac{d}{dt}\V{\phi} \in \GROUP{R}^{N_\phi},                                                                                                                                                                            \\
        \V{b}_{v} & = -\dot{\V{x}}^\intercal \left(\frac{\partial^2 \V{\phi}}{\partial \V{x}^2} \right) \dot{\V{x}} - 2 \frac{\partial^2 \V{\phi}}{\partial t \partial \V{x}} \dot{\V{x}} - \frac{d^2 \V{\phi}}{d t^2} \in \GROUP{R}^{N_\phi},
    \end{align}
\end{subequations}
where $N_x$ is the number of states of the system and $N_\phi$ the number of constraints imposed onto the system.

Integrating this system using common numerical integration methods (e.g. RK4) results in a small error that occurs after each integration step, caused by the fact that $\SET{S^3}$ is not closed for addition $\left(\Qx{q} + \Qx{q_{\delta}} \notin \SET{S^3}\right)$. The authors of~\cite{Xu2016} have shown how this error accumulates and alters the dynamics of the system. In practice, however, the error is relatively small under the condition that the time steps are sufficiently small, in which case~\eqref{eq:small_variation} holds. Furthermore, it is not required to re-normalize the quaternion after each iteration since the constraint drives the unity error to zero.

\section{Validation}
In the following section, the physical fidelity of the obtained constrained system model \eqref{eq:constrained-system}, is validated.

\subsection{Methodology}
Correctness of the model is shown by comparing the evolution of a dynamics simulation of an AM with a 1-DOF manipulator (see figure \ref{fig:validation-rotorcraft} using the constrained first order system \eqref{eq:constrained-system} and comparing them with the simulation as performed by the \textit{Bullet} physics engine \cite{PyBullet} using the same URDF model in both cases. The \textit{Bullet} simulation will thus be the reference model, called the \textit{Bullet model}. For clarity, the model corresponding to \eqref{eq:constrained-system} will be referred to as the \textit{Lagrange model}.

The simulation in both cases is performed at a fixed time step of $\SI{240}{\hertz}$. The simulation of the Lagrange model is using the Euler forward (first order) integration scheme, \textit{Bullet} physics uses the semi-implicit Euler method. As far as possible, equal conditions have been established for both simulations (e.g., the default linear and angular damping in \textit{Bullet} has been turned off for this test).

A short dynamics simulation of the modeled AM is carried out, during which positions, velocities, and propulsion forces are recorded. Given the inherently unstable nature of the AM, it is stabilized via a stabilizing (sliding mode) controller similar to \cite{Kim2013} around the equilibrium point. During the simulation, the first (and only) joint $\theta_{1}$ is driven towards the commanded reference position, following the pattern shown in figure \ref{fig:qerr-joints-misc} (top). The movement of that joint is thus acting as a disturbance onto the closed-loop system and stresses the various terms in~\eqref{eq:system-dynamics}. Caused by the accumulation of small errors at each simulation step, the two systems will deviate from each other eventually. Furthermore, the same control input will no longer stabilize both systems. This limits the time interval over which the dynamics of both systems are comparable. Here, this interval was identified to be around $\SI{4}{\second}$.

\begin{figure}
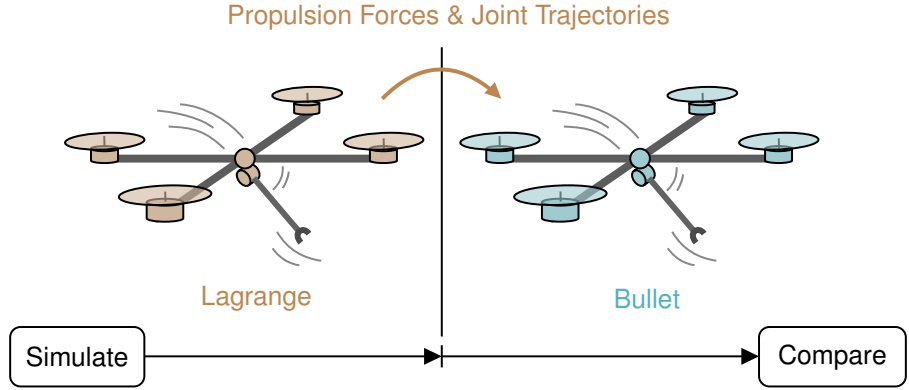

    \centering
    \include{figure_sim}
    \caption{The validation scenario. The model simulated using the model developed inhere (left), transfers the propulsion forces and joint trajectories to the (physically) identical model simulated with \textit{Bullet} (right). The two models are identical if they evolve dynamically close over a time $t$.}\label{fig:sim-vehicle}
\end{figure}

The recorded forces applied to the Lagrangian-model via the propulsion devices are then copied to the Bullet-model, which is thus pseudo-open loop. If both models are sufficiently close, the exact same forces should stabilize both models (within a reasonable time span). The joint torques on the other hand are not transferred due to numerical stability concerns, but since the joint is following the same trajectory, the generated disturbance is the same in both cases. Therefore, the Bullet-model tracks the joint velocity and position of the Lagrange-model via \textit{Bullet's} built-in \textit{joint position control} algorithm. The process is depicted in figure~\ref{fig:sim-vehicle}.

The hypothesis is thus, if both systems have the same intrinsic dynamics, the time response of both systems will be the same, resp. very close. Two systems will diverge eventually due to the accumulation of small errors over time e.g. caused by numerical inaccuracies, fundamental architectural differences, and different integration schemes.

Lastly, by performing a 'backflip' maneuver during which the simulated AM, momentarily points straight up resp. down, we can verify that our proposed model remains consistent where the model using Euler angles \cite{Kim2013} fails due to gimbal lock \cite{Hemingway2018}. To that end, the aforementioned 1-link AM is commanded to flip about its local Y-axis, thus pitching the platform by more than \SI{90}{\degree}.

\subsection{Results}
\begin{figure}
    \centering
    \begin{subfigure}[t]{.45\linewidth}
        \centering
        \includegraphics[width=\linewidth]{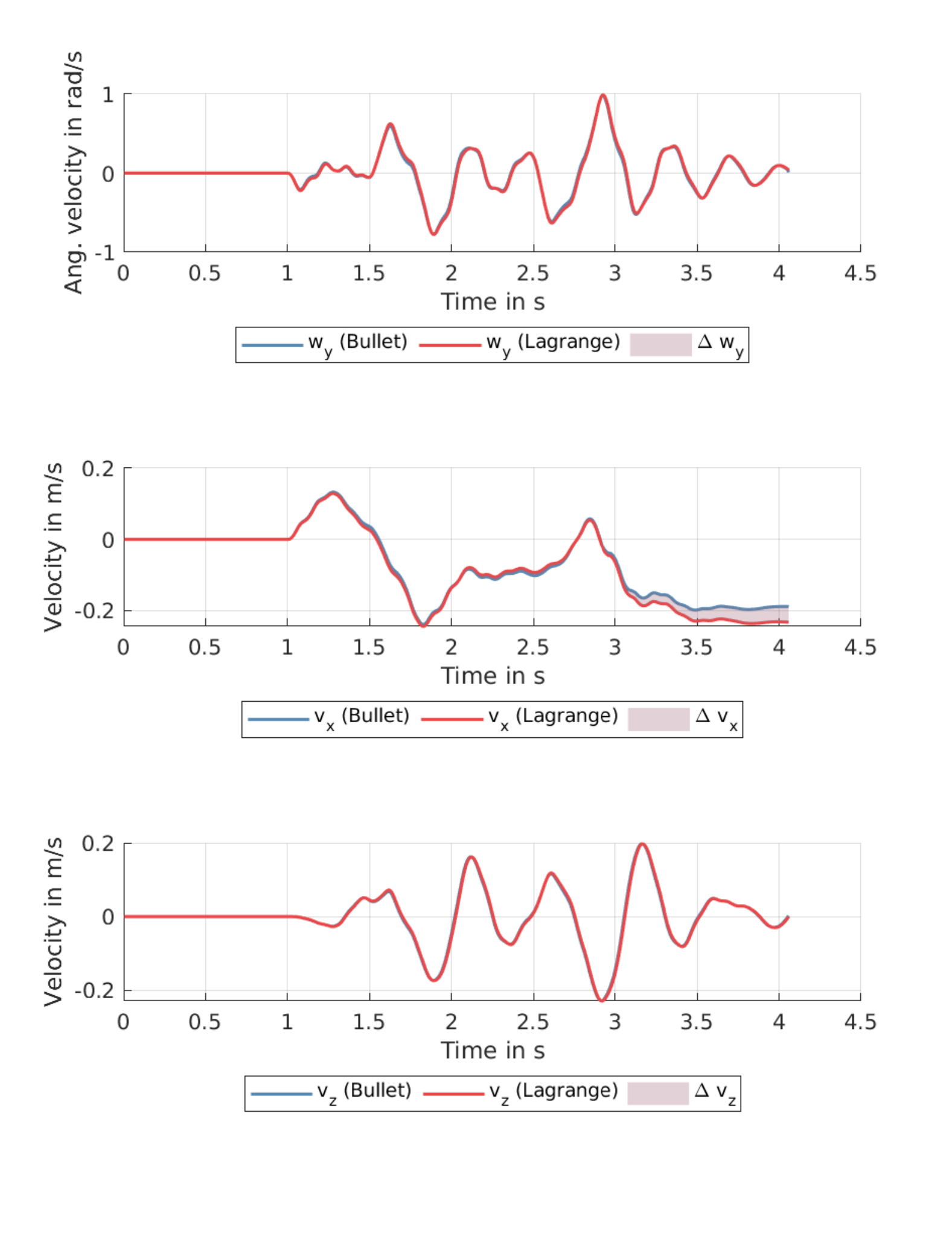}
        \caption{Angular velocity about the y-axis (top), velocity in y-direction (center), velocity in z-direction (bottom), for both the Bullet-model and the Lagrange-model as well as their mutual deviation.}\label{fig:dwy-dz-dy}
    \end{subfigure}%
    \hspace{0.5cm}
    \begin{subfigure}[t]{.45\linewidth}
        \centering
        \includegraphics[width=\linewidth]{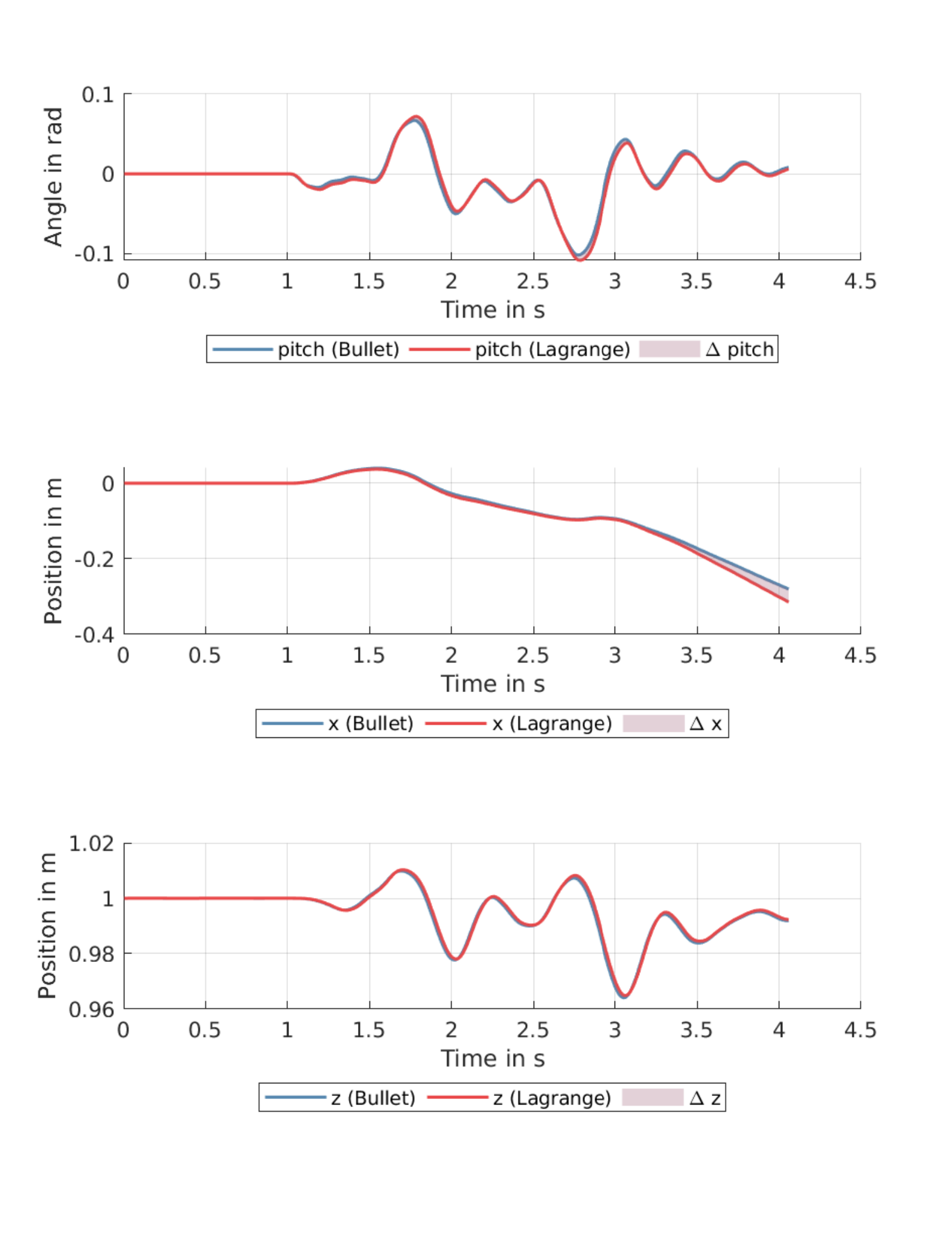}
        \caption{Angular position about the pitch axis (top), position in x direction (center), position in z-direction (bottom), for both the Bullet-model and the Lagrange-model as well as their mutual deviation.}\label{fig:pos-pitch-x-z}
    \end{subfigure}
    \caption{Validation results}\label{fig:validation1}
\end{figure}

\begin{figure}
    \centering
    \includegraphics[width=0.5\linewidth]{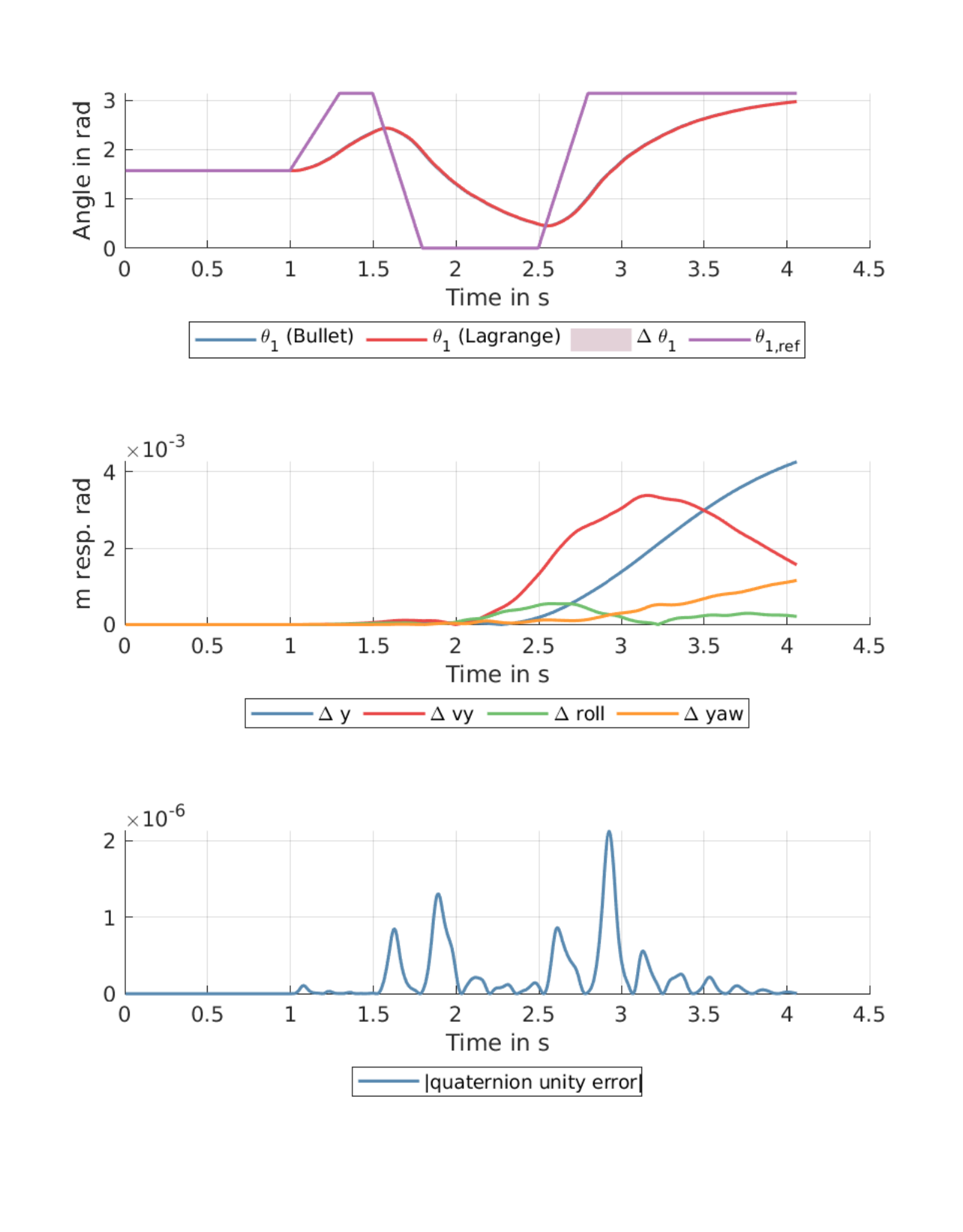}
    \caption{Joint positions (top) and the error on all remaining axis (center) of the AM. The absolute quaternion unity error $abs(\NORM{q}-1)$ (bottom) is driven to zero by the holonomic constraint.}\label{fig:qerr-joints-misc}
\end{figure}

\begin{figure}
    \begin{minipage}{0.49\linewidth}
        \centering
        \begin{tabular}{ll}
            \toprule
            Property            & Value
            \\\toprule
            UAV Configuration   & Quadcopter                                                       \\\midrule
            \#joints \& links   & $1$                                                              \\\midrule
            Joint Type          & Revolute (y axis)                                                \\\midrule
            Wheel base          & $\SI{1.51}{\meter}$                                              \\\midrule
            Mass (base)         & $\SI{6}{\kilo\gram}$                                             \\\midrule
            Inertia (base)      & $\left(0.48,0.48,0.95\right)\SI{}{\kilo\gram\meter\squared}$     \\\midrule
            Mass (link 1)       & $\SI{1}{\kilo\gram}$                                             \\\midrule
            Inertia (link 1)    & $\left(0.595,3.824,3.7\right)\SI{e-3}{\kilo\gram\meter\squared}$ \\\midrule
            CM (link 1)         & $\left(0.5, 0, 0\right)\SI{}{\meter}$                            \\\midrule
            $k_t$               & $\SI{2.165e-06}{\newton\per\minute\squared}$                     \\\midrule
            $k_p$               & $\SI{5.865e-08}{\newton\meter\per\minute\squared}$               \\\midrule
            $P$ (prop)          & $\SI{4500}{\per\minute}$                                         \\\midrule
            $P$ (joint)         & $\SI{16}{\newton\meter}$                                         \\\midrule
            $T$ (prop \& joint) & $\SI{0.2}{\second}$                                              \\\bottomrule
        \end{tabular}
    \end{minipage}\hfill
    \begin{minipage}{0.49\textwidth}
        \centering
        \includegraphics[width=1\columnwidth]{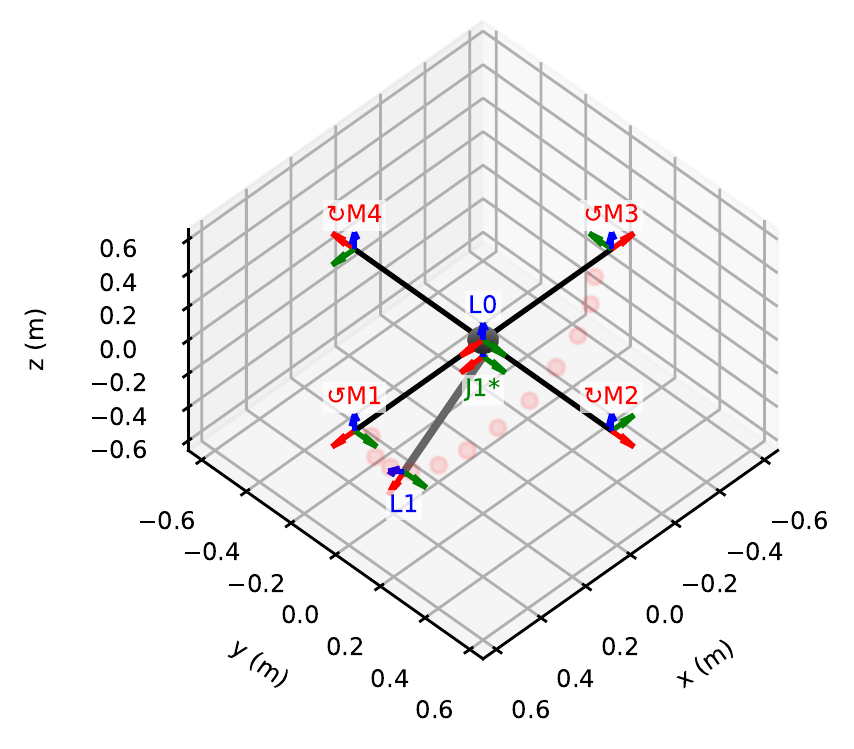}
    \end{minipage}
    \captionof{figure}{Physical properties of the Lagrange-model and the corresponding configuration of the quadcopter with a 1 DOF manipulator (rotating about its local y axis) used for the validation test. The range of motion of the manipulator is indicated by the red dots. The present manipulator configuration corresponds to $\V{\theta}=\left[\SI{0}{\degree}\right]$}\label{fig:validation-rotorcraft}
\end{figure}

The actuation of the first joint following the commanded pattern as shown in figure \ref{fig:qerr-joints-misc} (top), is creating a disturbance on the system. The manipulator moves in XZ plane (figure \ref{fig:validation-rotorcraft}), with the reaction torque acting on the Y axis (pitch) and the Coriolis force pushing the vehicle in X resp. Z direction. The resulting motion is shown in figures \ref{fig:validation1} and \ref{fig:qerr-joints-misc}.

The resulting velocities along the main axis of the disturbance are shown in figure \ref{fig:dwy-dz-dy}. The reaction torque creates a net torque around the Y axis, causing the system to accelerate, and turn about the Y axis. The same applies to the velocities in X and Z caused by the Coriolis forces initiated by the circular motion of the moving mass (the link here). The corresponding positions are shown in figure \ref{fig:pos-pitch-x-z}. Noticeably, is that the evolution of both models is extremely close, however, as can be seen in figure \ref{fig:dwy-dz-dy} (center), small errors do sum up and thus both systems deviate from each other eventually. As the states of both models deviate eventually, the control input will no longer stabilize both systems. A comparison thus only makes sense over a short interval ($\SI{4}{\second}$).

The yaw and roll position discrepancies are shown in figure \ref{fig:qerr-joints-misc}. Since the manipulator passes, during its movement, straight through the center of mass (in XZ plane), there should be no rotation about roll, which is the case here as it is measuring close to zero. The same reasoning applies to the yaw axis, except that also the propeller drag comes into play here. In Y direction, the AM also exhibits a net-zero movement as the manipulator moves in XZ plane.

The absolute quaternion unity error is shown in figure \ref{fig:qerr-joints-misc} (bottom). Since this is the fundamental assumption made at the start, it must deviate from unity as little as possible. The maximum error measured here was only $\SI{2.9e-6}{}$ indicating that the holonomic constraint solver pushes the quaternion norm to stay close to unity. It can also be seen that re-normalizing the quaternion after each iteration is unnecessary.

It can thus be concluded that our model is true to the reference model as simulated by the \textit{Bullet} realtime physics engine.

\begin{figure}
    \centering
    \includegraphics[width=0.5\columnwidth]{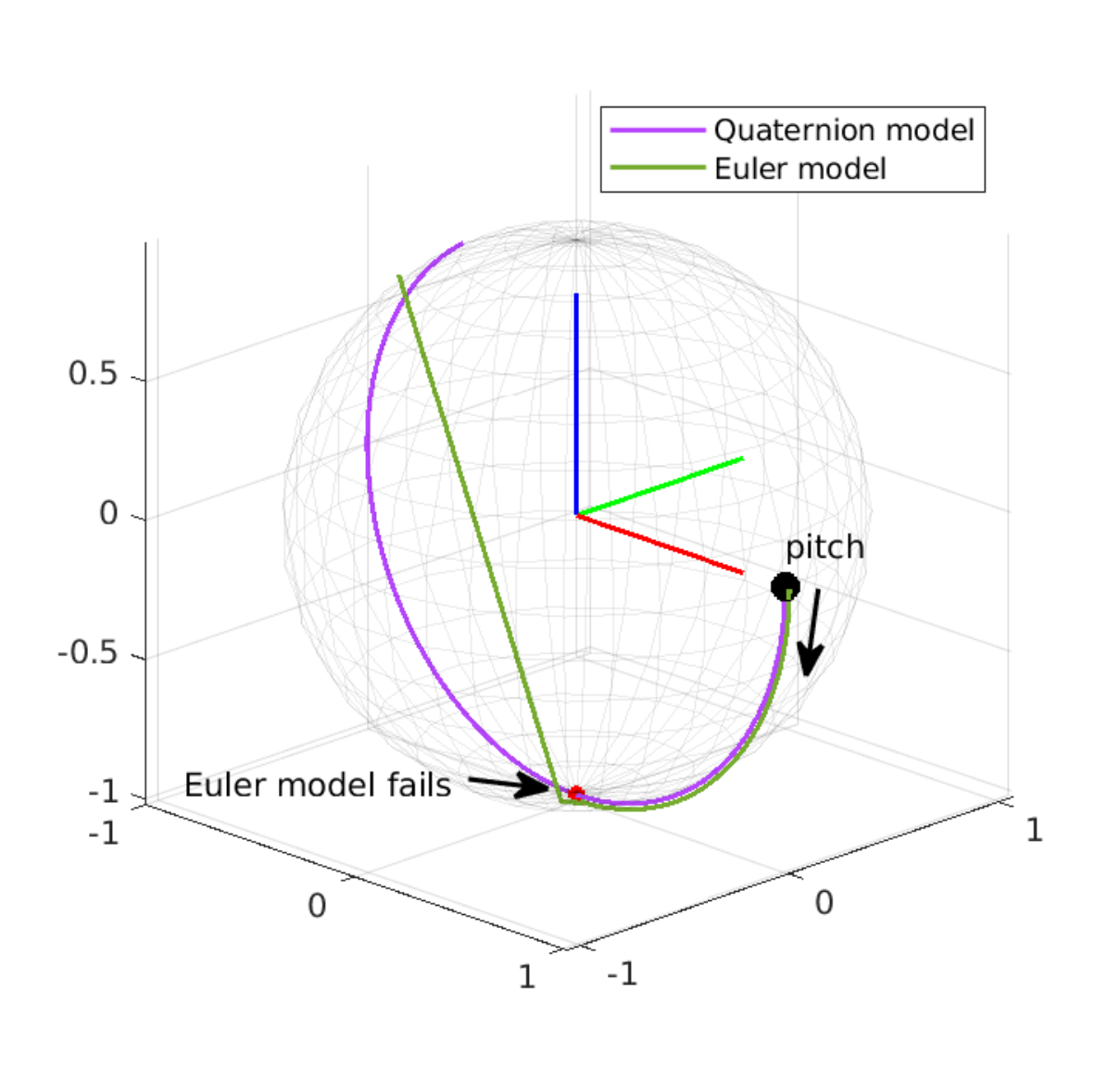}
    \caption{Quaternion parameterization avoids the singularities inherent to Euler parameterization. Here, the UAV is commanded to pitch (backflip), which drives the Euler model into gimbal lock as it approaches a pitch angle of $\SI{90}{\degree}$ close to the south pole, ultimately causing the motion to become unpredictable. The two trajectories represent the vehicle's x-axis plotted on a unit-sphere. The sudden jump on the Euler trajectory (green) is caused by gimbal lock.}
    \label{fig:quaterion_euler_motivation}
\end{figure}

Lastly, a backflip maneuver pitching the AM by more than $\SI{90}{\degree}$ is shown in figure \ref{fig:quaterion_euler_motivation}. The results show that the Euler model fails close to the south pole ($\SI{90}{\degree}$ mark) as it enters a gimbal lock condition, where the movement becomes unpredictable as the transformation matrix tying Euler rates to angular velocities becomes rank deficient. Consequently, the system states become 'NaN' - terminating the simulation. Our quaternion model has no such limitations and remains consistent.


\section{Control}
This section briefly demonstrates an application of the developed model in a control context applied to an AM consisting of quadcopter carrying a 2-DOF serial link manipulator. Furthermore, investigations are performed to analyze the realtime applicability of the control scheme with increasingly complex AM systems.

\subsection{Computed Torque Control}
\begin{figure}
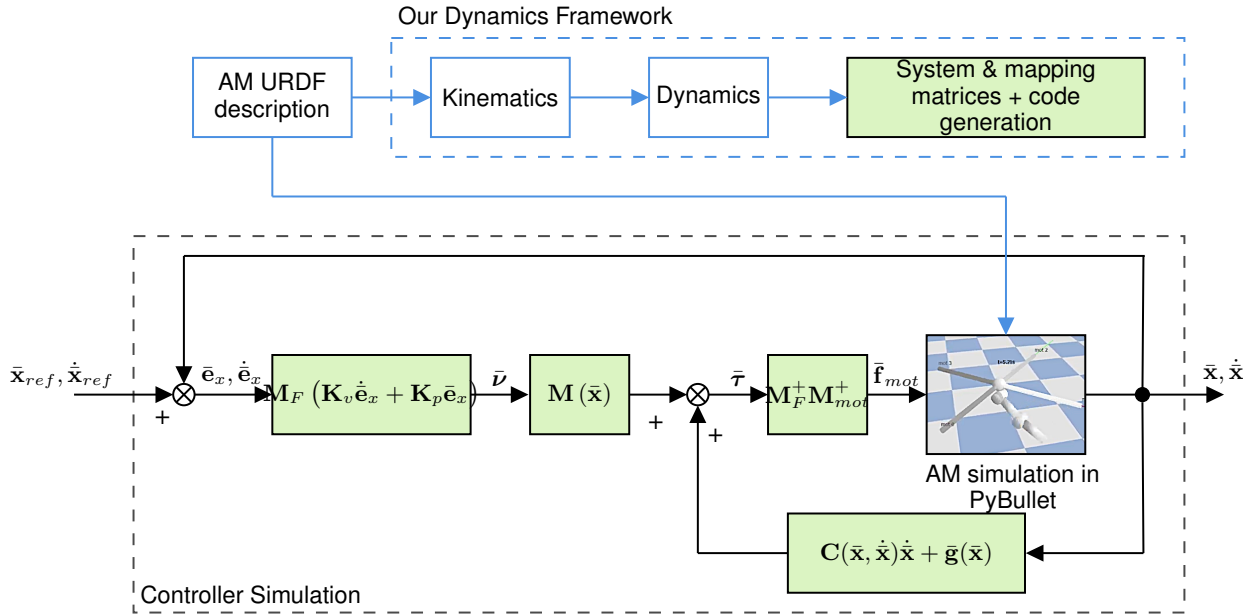

    \centering
    \include{figure_blockdiagram}
    \caption{Block diagram of the computed torque controller in the context of our dynamics framework. The controller's output, mapped to motor forces, is fed into the PyBullet simulation where it is applied to the simulated model of the AM in terms of local forces $\V{f}_{mot}$ at their corresponding locations. The positions and velocities of the simulated model are then fed back into the controller. The system matrices of the AM are generated by our dynamics framework based on a URDF description which also serves as the blueprint for the model inside PyBullet.}\label{fig:control-block-diagram}
\end{figure}

The controller is largely based on the computed torque controller presented in \cite{BrunoSiciliano2008} and depicted in figure \ref{fig:control-block-diagram}. The controller makes use of the model for feedback linearization and is, at its core, a PD controller with bias terms (known disturbances). This very simple structure was deliberately chosen for the subsequent tests as it does not hide model imperfections, contrary to the various state-of-the-art robust control schemes, which are largely preferable for real-world applications. Nonetheless, they also make heavy use of a dynamics model such as presented in this work.

Given the non-linear AM system
\begin{equation}
    \M{M}\ddot{\V{x}} + \M{C}\dot{\V{x}} + \V{g} = \V{\tau}, \label{eq:control-system}
\end{equation}
by deliberately choosing the control input
\begin{equation}
    \V{\tau}=\M{M}\V{\nu} + \M{C}\dot{\V{x}} + \V{g} \label{eq:control-input}
\end{equation}
that cancels the nonlinearities via the bias term $\M{C}\dot{\V{x}} + \V{g}$, the following system is obtained by inserting~\eqref{eq:control-input} in~\eqref{eq:control-system}:
\begin{equation}
    \ddot{\V{x}} = \V{\nu}. \label{eq:linearized-system}
\end{equation}
Using a PD-feedback law, the virtual control input $\V{\nu}$ can be defined by
\begin{equation}
    \V{\nu} = \M{M}_F \left( \ddot{\V{x}}_d + \M{K}_v  \dot{\V{e}}_x + \M{K}_p \V{e}_x \right), \label{eq:virtual-control-input}
\end{equation}
which, by inserting~\eqref{eq:virtual-control-input} in~\eqref{eq:linearized-system} yields the linear error dynamics
\begin{equation}
    \M{M}_F \left( \ddot{\V{e}}_x + \M{K}_v \dot{\V{e}}_x + \M{K}_p \V{e}_x \right) = \V{0}, \label{eq:error-dynamics}
\end{equation}
which is stable and guaranteed to converge by the right choice of the gain matrices $\M{K}_v$ and $\M{K}_p$ according to linear control theory. The mapping matrix $\M{M}_F$ \eqref{eq:force-mapping-matrix} maps the pseudo-forces of the controller into generalized coordinates.

The positional error vector $\V{e}$ is defined as
\begin{equation}
    \V{e} = \begin{bmatrix}
        {\Vf{W}{p}{W}{L_0}}_{,ref} - \Vf{W}{p}{W}{L_0} \\
        \V{q}_{v,err}                                  \\
        \V{\theta}_{ref} - \V{\theta}
    \end{bmatrix},
\end{equation}
where $\V{q}_{v, err}$ is the vector part of the error quaternion defined as $\Qx{q}_{err} = \Qx{q}^* \otimes \Qx{q}_{ref}$.

The velocity error is given by
\begin{equation}
    \dot{\V{e}} = \begin{bmatrix}
        {\Vfdot{W}{p}{W}{L_0}}_{,ref} - \Vfdot{W}{p}{W}{L_0} \\
        {\Vf{B}{\omega}{B}{B}}_{,ref} - \Vf{B}{\omega}{B}{B} \\
        \dot{\V{\theta}}_{ref} - \dot{\V{\theta}}
    \end{bmatrix}.
\end{equation}
The acceleration error term $\ddot{\V{e}}$ is not used as $\ddot{\V{x}}$ is difficult to obtain in practice (in good quality), thus $\ddot{\V{e}}=\V{0}$.

Lastly, the body forces to be applied by the controller are obtained by calculating
\begin{equation}
    \V{f}_b = \M{M}_F^+ \V{\tau},
\end{equation}
which can easily be mapped to motor forces using the mapping matrix \eqref{eq:body-force-to-motor-mapping}:
\begin{equation}
    \V{f}_{mot} = \M{M}_{mot}^+ \V{f}_b.
\end{equation}
The motor forces can then further be mapped to RPM via relation \eqref{eq:motor-thrust-rpm}.

It is clear that, on a real AM, the system matrices will often be based on estimations resp. approximations and will thus not exactly cancel the system's nonlinearities, which will degrade the performance of the controller. A common approach is thus to rely on the robustness of the controller (e.g.\cite{Heredia2014b}) or to make it adapt to miscalculated mechanical properties as shown in \cite{Kim2013}.

\subsection{Tests and Methodology}
A quadcopter equipped with a planar 2-DOF manipulator featuring two revolute joints is modeled in URDF and loaded into the realtime physics simulation \textit{Bullet}. The motor and joint forces are modeled as a first-order system \eqref{eq:motor-system} and applied locally as external forces resp. torques to the simulated body calculated with equation \eqref{eq:body-forces}. The system matrices ($\M{M}$, $\M{C}$, $\V{g}$) are generated from the same URDF description and exported to C code for performance reasons resp. to achieve close to bare metal performance for the subsequent performance benchmarks. The simulation and the controller are implemented in Python, making use of \textit{PyBullet} for the realtime physics simulation, \textit{numpy} for the linear algebra and \textit{cython} to wrap and utilize the generated code. The overall structure is as depicted in figure \ref{fig:control-block-diagram}.

\begin{figure}
    \begin{minipage}{0.49\linewidth}
        \centering
        \begin{tabular}{ll}
            \toprule
            Property            & Value                                                            \\\toprule
            UAV Configuration   & Hexacopter                                                       \\\midrule
            \#joints \& links   & $2$                                                              \\\midrule
            Wheel base          & $\SI{1.51}{\meter}$                                              \\\midrule
            Mass (base)         & $\SI{6}{\kilo\gram}$                                             \\\midrule
            Inertia (base)      & $\left(0.48,0.48,0.95\right)\SI{}{\kilo\gram\meter\squared}$     \\\midrule
            Mass (link 1\&2)    & $\SI{0.78}{\kilo\gram}$                                          \\\midrule
            Inertia (link 1\&2) & $\left(0.595,3.824,3.7\right)\SI{e-3}{\kilo\gram\meter\squared}$ \\\midrule
            CM (link 1\&2)      & $\left(0.22, 0, 0\right)\SI{}{\meter}$                           \\\midrule
            $T$ (joint)         & $\SI{0.05}{\second}$                                             \\\midrule
            $T$ (prop)          & $\SI{0.1}{\second}$                                              \\\midrule
            $P$ (prop)          & $\SI{4500}{\per\minute}$                                         \\\midrule
            $P$ (joint)         & $\SI{12}{\newton\meter}$                                         \\\midrule
            $k_t$               & $\SI{2.165e-06}{\newton\per\minute\squared}$                     \\\midrule
            $k_p$               & $\SI{5.865e-08}{\newton\meter\per\minute\squared}$               \\\bottomrule
        \end{tabular}
    \end{minipage}\hfill
    \begin{minipage}{0.49\textwidth}
        \centering
        \includegraphics[width=1\columnwidth]{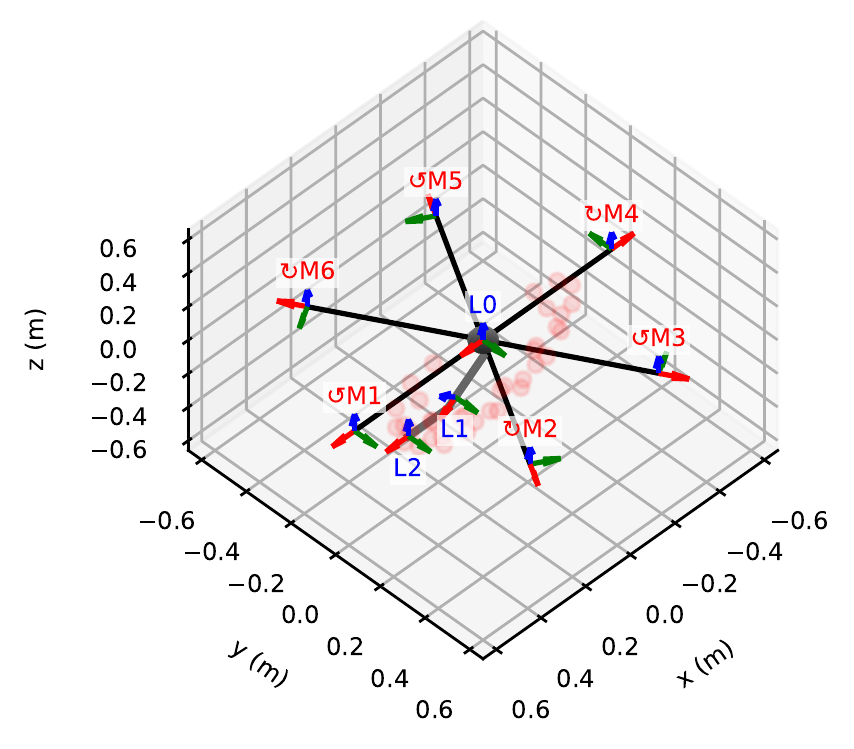}
    \end{minipage}
    \captionof{figure}{Physical properties and configuration of the controlled quadcopter equipped with a 2-DOF manipulator. The range of motion is indicated by the red dots. The present manipulator configuration corresponds to $\V{\theta}=\left[\SI{45}{\degree}, \SI{-45}{\degree}\right]$.}\label{fig:rotorcraft-control}
\end{figure}

The properties and the configuration of the AM is shown in figure \ref{fig:rotorcraft-control}, the gains of the controller were hand-tuned and read
\begin{align*}
    \M{K}_v & =\text{diag}\left(0,0,10,5,5,4,24,24\right),     \\
    \M{K}_p & =\text{diag}\left(0,0,30,40,40,30,60,120\right).
\end{align*}
The first two coefficients are zero for the velocity resp. position in $x$ and $y$ due to the lack of direct control authority over those quantities.

Two different tests are carried out. The first one analyzes the tracking performance of the computed torque controller in simulation over an interval of $\SI{20}{\second}$, performing several maneuvers whilst actuating the joints. The movement of the manipulator is a significant disturbance to the controlled system. Also, since the controller is in fact just a PD controller with a bias term, inconsistencies between the simulated model and the generated dynamics model would inevitably lead to an unstable system.

The second test is a series of benchmarks dealing with the realtime capability of the controller. The benchmarks are carried out on two different platforms that are representative for on-board resp. off-board control scenario. Herein, an \textit{AMD Ryzen 1600X} is used as a typical ground station, whilst a \textit{Raspberry Pi 4} was picked as a low-cost, low-power platform commonly used as a companion computer on drones to perform more substantial onboard calculations. Several AM systems from zero to 3 links were generated and analyzed. Prior to counting the number of operations on the left-hand side of \eqref{eq:system-dynamics} via \textit{SymPy}'s 'count\_ops' function, the expression is simplified with the 'expand' function. This operation is very costly but necessary to be able to get comparable results. For that reason, a maximum of 3 links was chosen for this test (no limitations are imposed by the framework itself). The generated C code is compiled with the \textit{LLVM Clang} compiler using the flags '-O1 -ffast-math -fno-math-errno'. The code is using the double-precision floating point format. The choice of using '-O1' (minimal optimizations) on the generated C code is motivated by the huge size of the generated code files (up to several megabytes). Performing only minimal optimizations helps with compile times and memory consumption during compilation. Tests with '-O2' only showed marginal improvements in terms of performance. All of the C++ code involving expensive linear algebra is compiled with standard release flags (i.e., '-O3'). For this series of tests, we got rid of all Python code and replaced the \textit{numpy} functionality with \textit{Eigen}, thus performing our benchmarks on a 'pure' C/C++ codebase.

The average time for one iteration of forward dynamics \eqref{eq:forw-dynamics} and inverse dynamics \eqref{eq:control-system} was measured and averaged over several runs. The forward dynamics is important for simulation, the inverse dynamics is what is fed into the controller in terms of known disturbances. The main difference between the two is the additional computational cost associated with the calculation of the inverse mass matrix and the constraint solver which comes into play for the forward dynamics.

Lastly, we compare the number of operations of the dynamics equation of a two-link AM using Euler and quaternion parameterization. Furthermore, we test different methods to obtain the dynamics equation from the general Lagrange equations. More precisely: \begin{enumerate*}
    \item The energy formulation from \cite{Wang2019}, \eqref{eq:mass-matrix-wang2019}, \eqref{eq:coriolis-vector-wang2019}, which results in a lumped Coriolis force vector.
    \item The factorization method used herein, involving the Christoffel symbols, \eqref{eq:system-mass-matrix}, \eqref{eq:system-coriolis-matrix}.
    \item A mix of the two methods, which also yields a lumped expression for the Coriolis term, \eqref{eq:system-mass-matrix}, \eqref{eq:coriolis-vector-wang2019}.
\end{enumerate*}

\subsection{Results}
\begin{figure}
    \centering
    \includegraphics[width=1\columnwidth]{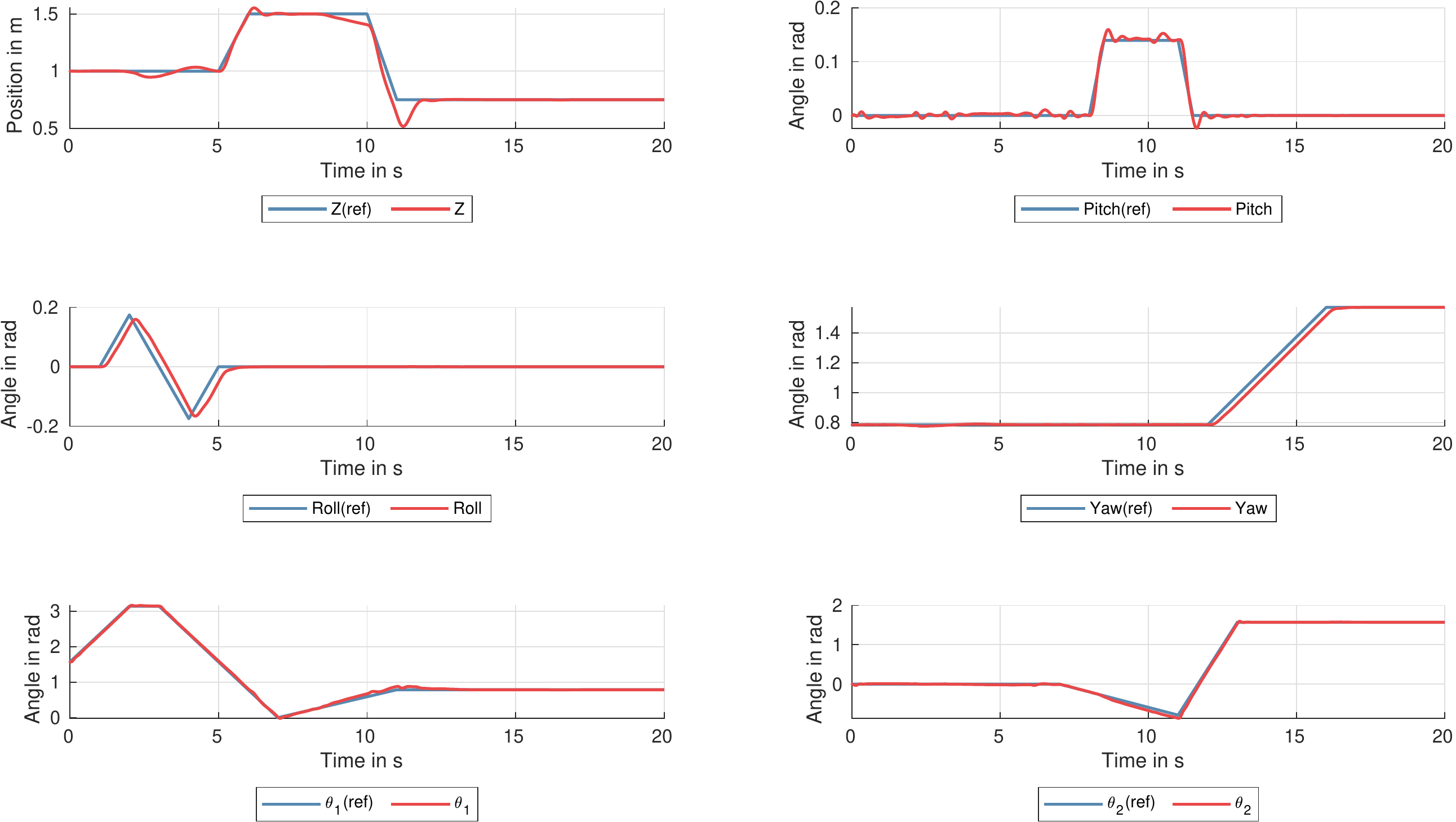}
    \caption{Simulation results of the computed torque controller applied to a PyBullet-simulated 2-link AM. Measured and reference signal for the position in $z$, the roll, pitch and yaw angles as well as the joint angles $\theta_1$ and $\theta_2$}\label{fig:control-results}
\end{figure}

The results of the simulation over a duration of $\SI{20}{\second}$ have been plotted in figure \ref{fig:control-results}. Reference trajectories (position and velocity) have been generated for the joint angles, the altitude, roll, pitch, and yaw angles, and fed to the controller. Notice that the controller tracks the reference quaternion obtained from the roll, pitch, and yaw angles. For convenience, those angles have also been plotted by converting the attitude quaternion back to roll, pitch and yaw.

The results in figure \ref{fig:control-results} show stable and satisfactory performance, indicating that the nonlinearities of the system are well compensated by the bias term. The steady-state error in $z$ direction between $\SI{8.5}{\second}$ and $\SI{10}{\second}$ is introduced by the pitch angle of the UAV, which leads to a loss in thrust in z-direction. This behavior is expected from a PD-controller (as is the care here). A robust control scheme can be used to fix this problem, or, in this case here, a simple addition of an integral term to the virtual control input would be sufficient.

Furthermore, as can be seen from figure \ref{fig:rotorcraft-control} resp. \ref{fig:validation-rotorcraft}, the time constants of the system have been tuned down from $\SI{0.2}{\second}$ to $\SI{0.1}{\second}$ for the propulsion resp to $\SI{0.05}{\second}$ for the joints. The time constants greatly impact the performance of the controller as they severely degrade its ability to cancel the nonlinearities via the bias term. The point could be made that AM systems with more but smaller propellers (thus with a smaller time constant) are preferable over systems with fewer but larger propellers.

The benchmark results concerning the computational time and realtime performance on the two model platforms (\textit{AMD Ryzen 1600X} and \textit{Raspberry Pi 4}) are shown in table \ref{tab:performance-results}. The results indicate that for the desktop CPU, the number of calculations hardly poses any problem with an average of just $\SI{66}{\micro\second}$ per forward dynamics iteration for a very typical UAV equipped with a 2-DOF manipulator, which then grows exponentially, reaching an average of $\SI{731}{\micro\second}$ for a UAV with a 3-DOF manipulator with a total of almost $6.6$ million arithmetic operations. As expected, the inverse dynamics are slightly faster, by the lack of matrix-inverse calculations, peaking at $\SI{706}{\micro\second}$ per iteration.

The Raspberry performs noticeably worse, reaching up to $\SI{4.93}{\milli\second}$ per forward dynamics iteration in the 3-DOF scenario and only slightly better at $\SI{4.89}{\milli\second}$ per inverse dynamics iteration. Whether or not this is acceptable for realtime control heavily depends on the application. Typically, the inner loop of commercial autopilots runs at $\SI{500}{\hertz}$, which would limit the \textit{Raspberry Pi} to a 2-link configuration.

From table \ref{tab:solver-compare} it becomes clear that the computational cost increase from choosing quaternion parameterization over Euler angles is substantial. Across the board, the number of operations roughly doubled, passing from Euler angles to quaternion parameterization. This is at least partially due to the larger state space and, thus all system matrices being enlarged by one over the Euler model. The numbers also show that the method itself has a big impact on the complexity of the resulting dynamics expression. We achieved the best results with the "mixed"-method,  \eqref{eq:system-mass-matrix}, \eqref{eq:coriolis-vector-wang2019} for the quaternion model. The "energy"-method, \eqref{eq:mass-matrix-wang2019}, \eqref{eq:coriolis-vector-wang2019}, delivered the best results for the Euler model. In all cases, a lumped expression for the Coriolis forces is preferable. Calculating the Christoffel symbols, \eqref{eq:system-mass-matrix}, \eqref{eq:system-coriolis-matrix}, always resulted in the largest expression. The presented methods are also vastly different in terms of the time it takes for the model to generate. Here, the mixed method was the fastest in all cases, while the energy method was the slowest. This can be attributed to the number of derivatives calculations involved in that method.

\begin{table}
    \centering
    \caption{Computational performance and number of arithmetic operations of several configurations on two typical compute platforms using equations \eqref{eq:system-mass-matrix}, \eqref{eq:system-coriolis-matrix}.}\label{tab:performance-results}
    \begin{tabular}{lrrrrr}
        \toprule
                      &                & \multicolumn{2}{c}{Ryzen 1600X} & \multicolumn{2}{c}{Raspberry Pi 4}                                                       \\
        Configuration & Operations     & Forw. dynamics                   & Inv. dynamics                     & Forw. dynamics            & Inv. dynamics           \\\toprule
        UAV           & \SI{5911}{}    & \SI{2}{\micro\second}           & \SI{0.6}{\micro\second}            & \SI{10}{\micro\second}   & \SI{2}{\micro\second}    \\\midrule
        UAV+1 Links   & \SI{130328}{}  & \SI{11}{\micro\second}          & \SI{9}{\micro\second}              & \SI{60}{\micro\second}   & \SI{46}{\micro\second}   \\\midrule
        UAV+2 Links   & \SI{1041911}{} & \SI{66}{\micro\second}          & \SI{63}{\micro\second}             & \SI{1031}{\micro\second} & \SI{991}{\micro\second}  \\\midrule
        UAV+3 Links   & \SI{6549828}{} & \SI{731}{\micro\second}         & \SI{706}{\micro\second}            & \SI{4925}{\micro\second} & \SI{4886}{\micro\second} \\\bottomrule
    \end{tabular}
\end{table}

\begin{table}
    \centering
    \caption{Comparison of the computational cost for quaternion and Euler angle parameterized models for a 2-link AM. Christoffel method uses \eqref{eq:system-mass-matrix}, \eqref{eq:system-coriolis-matrix}, energy method uses \eqref{eq:mass-matrix-wang2019}, \eqref{eq:coriolis-vector-wang2019} and the mixed method uses \eqref{eq:system-mass-matrix}, \eqref{eq:coriolis-vector-wang2019}.}\label{tab:solver-compare}
    \begin{tabular}{llcrrrr}
        \toprule
        Model      & Method      & Lumped & Operations           & Rel. Difference      & Abs. Difference      & Gen. Time                \\\midrule
        Quaternion & Christoffel & no     & \SI{1041911}{}       & baseline             & baseline             & \SI{760}{\second}        \\
                   & Energy      & yes    & \SI{701721}{}        & \SI{-32.7}{\percent} & \SI{-32.7}{\percent} & \SI{1146}{\second}       \\
                   & Mixed       & yes    & \best{\SI{698908}{}} & \SI{-32.9}{\percent} & \SI{-32.9}{\percent} & \best{\SI{613}{\second}} \\\midrule
        Euler      & Christoffel & no     & \SI{516078}{}        & baseline             & \SI{-50.5}{\percent} & \SI{284}{\second}        \\
                   & Mixed       & yes    & \SI{349888}{}        & \SI{-32.2}{\percent} & \SI{-66.4}{\percent} & \best{\SI{215}{\second}} \\
                   & Energy      & yes    & \best{\SI{192813}{}} & \SI{-62.6}{\percent} & \SI{-81.5}{\percent} & \SI{312}{\second}        \\\bottomrule
    \end{tabular}
\end{table}

\section{Conclusion}
The kinematic equations for the multi-body system formed by an AM (i.e., UAV and a serial link manipulator attached thereto), have been derived starting with a general URDF description of the system. The complete closed-form, singularity-free dynamics of an AM has been derived in detail via the Euler-Lagrange equations using quaternion parametrization for the rotational degrees of freedom of the floating base. The dynamics equations of the system were then factorized into the canonical form for mechanical systems.

The obtained dynamics model was numerically validated against its counterpart (simulated by the \textit{Bullet} physics engine), showing adequate performance over a reasonable time span. The results also show that the unity constraint of the quaternion stays satisfied over the course of the simulation by enforcing it via a general-purpose holonomic constraint solver.

The relation between the generalized forces and the body forces has been established whilst providing detailed insights on how those forces are calculated resp. modeled within an AM context.

And lastly, as an application, the stabilization of an AM using a computed torque controller making use of the developed dynamics model was shown, indicating satisfactory tracking performance in most cases. A robust control approach is, however, necessary to compensate for the loss of vertical thrust due to the inclination of the flying platform. Our data shows that, although the complexity of the closed-form solution grows exponentially, the proposed solution is generally fast enough for realtime applications of typical 2-link AM systems on low-power embedded platforms. Our results also show that the computational cost increase of our quaternion model compared to the common Euler model is quite substantial. The selected method to extract the dynamics equations from the general Lagrange equations shows to have a significant impact on the complexity of the resulting expression.

The current implementation makes use of a general-purpose holonomic constraint solver to keep the quaternion at unit length, which currently prevents it from being used in a model predictive control framework. Future work will deal with this issue by eliminating the unity constraint and, thus, the need for the constraint solver.

\newpage
\appendix
\section{Quaternion Fundamentals}\label{sec:quaternion-fundamentals}
This appendix briefly introduces quaternions with a focus on unit quaternions, which play a fundamental role in representing singularity-free orientations in three-dimensional space. Its most relevant aspects are summarized here. For further details the reader is referred to~\cite{Graf2008},~\cite{Moller2012},~\cite{Sola2017} and~\cite{Sola2018}.

A quaternion is an expression defined by a set of four coefficients $q_w, q_x, q_y, q_z \in \GROUP{R}$, and three symbols $i, j, k$,
\begin{equation}
    \Qx{q} = q_w + q_x i + q_y j + q_z k \in \GROUP{H},
\end{equation}
which can be put in the more convenient form of
\begin{equation}
    \Qx{q} = \left(q_w, \V{q}_v\right) \in \GROUP{H},
\end{equation}
where $q_w\in\GROUP{R}$ is the scalar part and $\V{q}_v=\left(q_i, q_j, q_k\right)\in\GROUP{R}^3$ is the vector part containing the three imaginary coefficients.

The quaternion group $\GROUP{H}$ is endowed with the non-commutative quaternion product defined as
\begin{equation}
    \Qx{q} \otimes \Qx{p} = \left(
    q_w p_w - \V{q}_v^\intercal \cdot \V{p}_v,
    q_w \V{p}_v + p_w \V{q}_v + \V{q}_v \times \V{p}_v
    \right). \label{eq:quaternion-product}
\end{equation}

The norm of a quaternion is defined as
\begin{equation}
    \NORM{\Qx{q}} = \sqrt{q_w^2 + \V{q}_v^\intercal \V{q}_v}.
\end{equation}
In the context of mechanics, quaternions describe both a (uniform) scaling and a rotation~\cite{Moller2012}. A quaternion $\Qx{q}$ with $\NORM{\Qx{q}}=1$, element of $\SET{S}^3$, is called a \textit{unit quaternion}, parameterizing a 4-dimensional unit-sphere, and, contrary to a general quaternion in $\GROUP{H}$, solemnly describe the orientation of a rigid body. $\SET{S}^3$ is a double cover of $\SET{SO}(3)$ meaning that $\Qx{q}$ and $-\Qx{q}$ characterize the same orientation.

The inverse of a quaternion is defined as
\begin{equation}
    \Qx{q}^{-1} = \frac{1}{\NORM{\Qx{q}}^2} \left(q_w, -\V{q}_v\right).
\end{equation}

The conjugate of a quaternion is obtained by negating its vector part
\begin{equation}
    \Qx{q}^* = \left(q_w, -\V{q}_v\right),
\end{equation}
and is equal to the inverse in case $\NORM{\Qx{q}}=1$.

The quaternion identity rotation is defined as
\begin{equation}
    \Qx{1} = \Qx{q} \otimes \Qx{q}^* = \left(1, \V{0}\right), \Qx{q}\in\SET{S}^3.
\end{equation}

Unit quaternions can directly be obtained from axis-angle notation (the equivalent of the Euler notation in quaternion space) s.t.:
\begin{equation}
    e^{\V{u}\theta} = \left(\cos{\frac{\theta}{2}}, \V{u}\sin{\frac{\theta}{2}}\right),
    \label{eq:axis-angle-notation}
\end{equation}
where $\V{u}$ represents the (normalized) axis of rotation and $\theta$ the angle of rotation.

A quaternion is called pure if its scalar part is zero:
\begin{equation}
    \Qx{q} = \left(0, \V{q}_v\right) \in \GROUP{H}_p \cong \GROUP{R}^4.
\end{equation}

A vector $\V{u} \in \GROUP{R}^3$ defined in the body frame can be rotated into the world frame by $\Qx{q} \in \SET{S}^3$ using the double quaternion product:
\begin{equation}
    \left(0, \V{u}'\right) = \Qx{q} \otimes \left(0, \V{u}\right) \otimes \Qx{q}^*\in \GROUP{H}_p.
    \label{eq:sandwich-product}
\end{equation}
The vectors $\V{u}$ and $\V{u}'$ have the same magnitude if and only if $\NORM{\Qx{q}}=1$. Otherwise, the length of $\V{u}$ is scaled by the norm of $\Qx{q}$.

Composition of two quaternions $\Qx{q}_1, \Qx{q}_2 \in \SET{S}^3$ is similar to the composition of rotation matrices:
\begin{equation}
    \Qx{q}_{12} = \Qx{q}_1 \otimes \Qx{q}_2.
\end{equation}
The quaternion product is linear and as such it can be expressed as a matrix-vector product:
\begin{align}
    \Qx{q}_{12} & = \Qx{q}_1 \otimes \Qx{q}_2 \\
                & = \Q{q_1}{L} \Qx{q}_2       \\
                & = \Q{q_2}{R} \Qx{q}_1,
\end{align}
with the operators
\begin{align}
    \Q{q}{L} & = q_w \M{1}_4 + \begin{bmatrix}
        0       & -\V{q}_v^\intercal \\
        \V{q}_v & \CROSS{\V{q}_v}
    \end{bmatrix} \label{eq:ql}
\end{align}
and
\begin{align}
    \Q{q}{R} & = q_w \M{1}_4 + \begin{bmatrix}
        0       & -\V{q}_v^\intercal             \\
        \V{q}_v & -\CROSS{\V{q}_v} \label{eq:qr}
    \end{bmatrix},
\end{align}
wherein $\CROSS{.}: \GROUP{R}^3 \rightarrow \GROUP{R}^{3\times3}$ is the cross product operator
\begin{align}
    \CROSS{\V{a}} = \begin{bmatrix}
        0    & -a_z & a_y  \\
        a_z  & 0    & -a_x \\
        -a_y & a_x  & 0
    \end{bmatrix}. \label{eq:cross-product}
\end{align}
Applying~\eqref{eq:ql} and~\eqref{eq:qr} to~\eqref{eq:sandwich-product} yields the rotation of a vector in matrix notation:
\begin{align}
    \begin{bmatrix}
        0 \\ \V{u'}
    \end{bmatrix} = \Q{q}{L}\Q{q^*}{R} \begin{bmatrix}
        0 \\ \V{u}
    \end{bmatrix}
\end{align}
The matrix equivalent of~\eqref{eq:axis-angle-notation} is given by the Rodrigues rotation formula $R\left(\theta,\V{u}\right): \GROUP{R} \times \GROUP{R}^3 \rightarrow \SET{SO(3)}$, which represents a rotation of an angle $\theta$ around an arbitrary axis $\V{u}$ and is define as
\begin{equation}
    R\left(\theta,\V{u}\right) = \M{1}_3+\sin{\theta}\CROSS{\V{u}}+\left(1-\cos{\theta}\right)\CROSS{\V{u}}^2.
    \label{eq:rodrigues}
\end{equation}
Within this work, the quaternion is always assumed to be unit length.

\bibliographystyle{unsrt}
\bibliography{mybib}
\end{document}